\documentclass[letterpaper, 10 pt, conference]{ieeeconf}  

\IEEEoverridecommandlockouts                              

\usepackage{times}
\usepackage{multicol}
\usepackage[bookmarks=true]{hyperref}
\usepackage{graphicx}
\usepackage{amsbsy}
\usepackage{amsmath}
\usepackage{amssymb}
\usepackage{multirow}
\usepackage{caption}
\usepackage{booktabs}

\usepackage[subrefformat=simple, labelformat=simple]{subcaption}
\usepackage{floatrow}

\usepackage{textcomp}
\usepackage{algorithm}
\usepackage{algpseudocode}
\usepackage{color}
\usepackage{xspace}
\usepackage{enumerate}

\title{\LARGE \bf
Fast Robust Monocular Depth Estimation for Obstacle Detection with Fully Convolutional Networks
}

\author{Michele Mancini$^{1}$, Gabriele Costante$^{1}$, Paolo Valigi$^{1}$ and Thomas A. Ciarfuglia$^{1}$
\thanks{*We gratefully acknowledge the support of NVIDIA Corporation with the donation of the Tesla K40 GPU used for this research.}
\thanks{$^{1}$All the authors are with the Department of Engineering, University of Perugia, via Duranti 93, Perugia Italy}
\thanks{{\tt\footnotesize \{michele.mancini,gabriele.costante, thomas.ciarfuglia, paolo.valigi\}@unipg.it}}
}

\def\LatexSettings{./settings}  
\def\include{./include}

\input{\LatexSettings/def_macros.tex} 

\begin{document}

\maketitle
\thispagestyle{empty}
\pagestyle{empty}

\begin{abstract}
Obstacle Detection is a central problem for any robotic system, and critical for autonomous systems that travel at high speeds in unpredictable environment. This is often achieved through scene depth estimation, by various means. When fast motion is considered, the detection range must be longer enough to allow for safe avoidance and path planning. Current solutions often make assumption on the motion of the vehicle that limit their applicability, or work at very limited ranges due to intrinsic constraints. We propose a novel appearance-based Object Detection system that is able to detect obstacles at very long range and at a very high speed ($\mathbf{\sim300Hz}$), without making assumptions on the type of motion. We achieve these results using a Deep Neural Network approach trained on real and synthetic images and trading some depth accuracy for fast, robust and consistent operation. We show how photo-realistic synthetic images are able to solve the problem of training set dimension and variety typical of machine learning approaches, and how our system is robust to massive blurring of test images. 
\end{abstract}

\section{Introduction}
Obstacle Detection (OD) is a challenging and relevant capability for any autonomous robotic system required to operate in real world scenarios, for safe path planning tasks and reaction to unexpected situations. Obstacle pose estimation must be fast enough to allow robot control system to react and perform required corrections. Since higher robot speeds require longer range detection to timely react, OD in automotive and autonomous aerial vehicle applications is particularly challenging.
Obstacle definition changes according to the specific application. In automotive and ground-based robotic applications an obstacle is usually any vertical object raising from the ground, such as cars, pedestrian, traffic lights poles, garbage bins, trees etc. 
When Micro aerial Vehicles (MAVs) are considered, some other assumptions are required. For example horizontal structures, such as tree branches and overpasses, signs become relevant obstacles, since robot motion is no more constrained to a well defined street environment. In these cases the OD system has to be able to detect any physical object present in the scene. 

\begin{figure}[ht!]
	\includegraphics[scale=0.2]{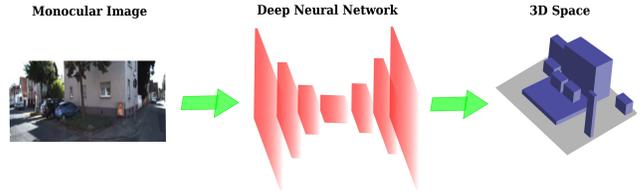} 
	\caption{\small We propose a fully convolutional network fed with both images and optical flows to obtain fast and robust depth estimation, with a robotic applications-oriented design. \label{fig:abstract}}
\end{figure}
The main techniques to address the OD problem are based on visual stereo systems. However, such systems are limited in detection range and accuracy by camera set-up and baselines \cite{davies2004machine}, \cite{pinggera2014know}, which in turn pose a limit on maximum speed, and this is a tough constraint both in automotive and MAV applications.
To overcome this limitations some systems exploited geometric knowledge about obstacles relationships with ground plane and assuming a limitation in the degrees of freedom of the vehicle’s movement, allowing long range obstacle detection up to 200 meters \cite{pinggera2015high} or a real-time construction of a raw 3D obstacle mapping \cite{cordts2014object}, \cite{benenson2011stixels}, \cite{pillai2015high}. Unfortunately, these methods can't work in application where their geometric assumptions are violated and the robot does not operate at ground level.

Monocular based vision detection systems have been proposed to bypass both stereo vision limitations and geometric assumptions. Since monocular vision does not allow accurate and robust distance geometric measurement, often machine learning based solutions have been proposed \cite{michels2005high}, \cite{dey2014vision}. Since learning methods are limited by the training set samples and these methods have been trained using datasets with ground truths collected through stereo vision or laser rigs, these solutions still have limitations on range and accuracy as stereo systems.

To develop an OD system that is capable of detecting obstacles at high speed, allowing fast motion without geometrical assumptions, we propose an hybrid monocular approach that trades some detection accuracy for speed and general applicability. We decided to use monocular images to be able to apply the method on small or micro aerial vehicles that are able to move up to speeds of 10-20 m/s, for which a stereo approach would not be viable. In addition, we address the problem using Deep Neural Networks (DNNs) to learn an algorithm that is accurate and fast enough to allow fast reaction to unexpected obstacles on the vehicle path. To solve the limitations of machine learning approaches, namely the lack of data and generality of the solution, we extend the dataset with artificial sequences created using a state-of-the-art graphic engine capable of producing photo-realistic outdoor environments. This allows us to add an arbitrary number of sequences with perfect ground truth at very long distances (200m), that would not have been possible to collect with a laser or stereo based ground truth system. Through our experiments we show that our algorithm is capable of doing fast estimation of depth with an accuracy that is sufficient for motion planning and that the learning on simulated photo-realistic environments is a viable way to extend datasets on robot vision problems. 

This work is focused on depth estimation for obstacle perception and does not assess planning and control strategies to achieve effective obstacle avoidance. These aspects will be considered in future works.  


\section{Related Work} \label{sec:related_work}

Most of traditional vision-based obstacle avoidance works rely on stereo vision. The most trivial solutions are based on finding disparities between the two matched images, compute point clouds and apply heuristics to detect obstacles. This methodology suffers from range limitations, produces sparse maps and may be not robust to pixel matching errors \cite{goldberg2002stereo}. Many of these methods are based on \textit{v-disparity} computation. Labayrade \etal \cite{labayrade2002real}, using a planarity assumption on stereo cameras, formulates a more robust analytical ground-plane estimation method based on \textit{v-disparity} computation. Benenson \etal \cite{benenson2011stixels} use v-disparity and \textit{u-disparity} to generate at high rate a fast obstacle representation on 3D space, while Harakeh \etal \cite{harakeh2015ground} build a probability field based on v-disparity to get a precise ground segmentation and occupancy grid of the scene. Pinggera \etal \cite{pinggera2015high} improve range and accuracy detection using stereo vision to compute local ground normal as a statistical hypothesis testing problem, getting detection range up to 200 meters. Pillai \etal \cite{pillai2015high} propose a tunable and scalable stereo reconstruction algorithm which allows scene depth comprehension with very high frame rates, which may be usable for real time obstacle detection purposes. 
The main issue of these methods is that they make geometric assumptions, requiring planarity between stereo images. Also, methods such as \cite{labayrade2002real}, \cite{benenson2011stixels}, \cite{harakeh2015ground} and \cite{pinggera2015high} utilize ground model to position obstacles on 3D space, so obstacles posed over the ground won't be detected by these methods. \\ 
Stereo vision has also been used as a data acquisition method for machine-learnind approaches. Hadsell \etal \cite{hadsell2009learning} use a stereo rig to assign labels to close-range obstacles, detecting them using geometric techinques cited above. Features are extracted from image patches containing those obstacles through an offline-trained convolutional autoencoder. A classifier is trained using these features and obtained labels as reference. Distant obstacles are then detected by the online trained classifier. Ball \etal \cite{ball2016vision} apply a similar approach optimized for agricultural applications, based on a novelty-based obstacle detector. 
To overcome stereo methods geometric constraints, monocular vision-based methods have been proposed. Mori \etal \cite{mori2013first} extract SURF features from monocular images and use template matching to detect frontal obstacles from their change in relative size between consecutive frames. This method makes no geometric assumption on the scene, but it has no capability to detect lateral obstacles and has limited range, which makes it unsuitable for high speed operations. Optical flow based obstacle detection has been explored in \cite{beyeler2009vision}, but it tends to be noisy in images with far away backgrounds, where optical flow tends to assume values close to zero. Day \etal \cite{dey2014vision} implement a imitation learning based reactive MAV controller based on monocular images, training a non-linear regressor to detect obstacles distances from features extracted from several patches of the image, as optical flow, histogram of oriented gradients, Radon transform, Laws' Masks and structure tensors. Ground truth for obstacles is obtained through a stereo rig. Being a machine learning based algorithm, it cannot perform better that the hardware used for training, so detection capability are limited by stereo vision weaknesses. 

In the set of monocular methods, we also consider depth map estimators. These methods solve a different problem, as they try to find an accurate 3D reconstruction of the scene, but we use them as benchmarks for our methods. Michels \cite{michels2005high} implements a reinforcement learning based 3D model generator with real-time capability. It relies on horizontal alignment of the images and does not generalize in less controlled settings. Eigen \etal \cite{eigen2014depth} develop a deep learning based architecture for single image 3D reconstruction trained, on different experiments, on NYUDepth dataset \cite{silberman2012indoor} and KITTI dataset \cite{Geiger12}, obtaining state-of-the-art performance in terms of depth estimation accuracy. Although, for robotic application, we're not interested into obtaining state-of-the-art precision as it would require higher computational costs and it would be even unnecessary for our purposes, we consider these methods as reference for estimation performance.

We share with some of these methods the learning approach based on Deep Neural Networks (DNNs), but we fetch them not only with monocular images, but also with the optical flow of consecutive frames. Our architecture is inspired by recent works in semantic segmentation and optical flow estimation (\cite{long2015fully},  \cite{noh2015learning}, \cite{badrinarayanan2015segnet}, \cite{fischer2015flownet}), where Fully Convolutional Networks (FCNs) have been trained to make pixel-wise estimations, obtaining outputs of the same size of the input image. Differently from standard Convolutional Neural Network approaches, FCNs do not make use of fully connected layers, which account for most of the parameters of the network (\eg on VGG-16 \cite{Simonyan14c} architecture fully connected layers parameters are about 120M, out of the 134M parameters describing the whole network), and for this reason they improve training speed and reduce the amount of data required to train the deep network \cite{badrinarayanan2015segnet}. In addition, since convolution operations can be strongly optimized on GPU, these networks can generate estimates with very high frame rates. 
\section{Network Structure} \label{sec:ns}

\begin{figure*}[ht|]
	\centering
	\includegraphics[scale=0.35]{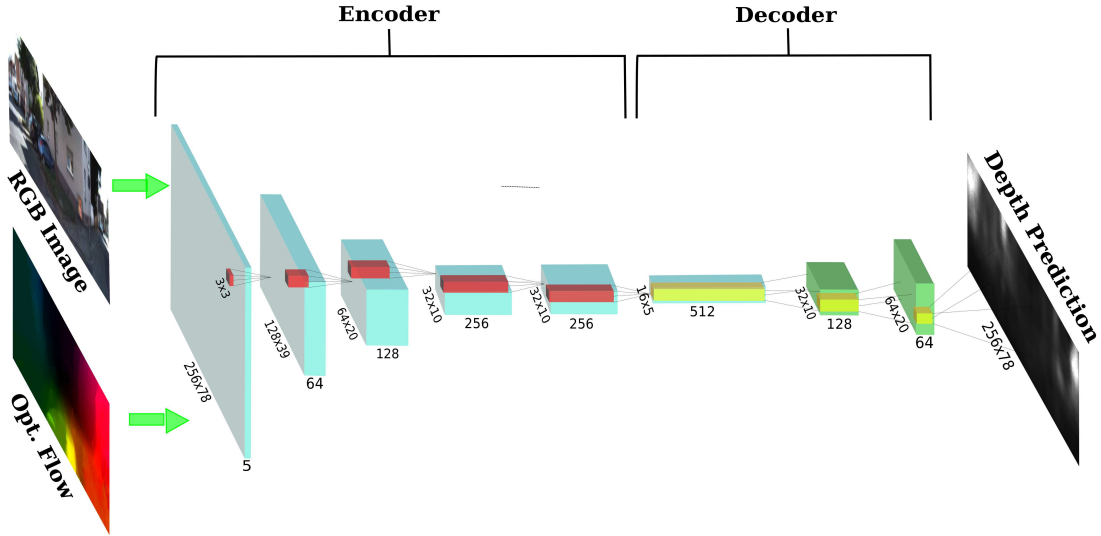} 
	\caption{\small Network architecture. Blue boxes: Encoder feature maps. Green boxes: Decoder feature maps. Convolutional filters are reported in red, deconvolutional filters in yellow.\label{fig:network}}
\end{figure*}

For the network structure we propose an \textit{encoder-decoder} architecture, similar to \cite{fischer2015flownet}, \cite{badrinarayanan2015segnet} and \cite{noh2015learning}. Since the problem we tackle is depth estimation for Obstacle Detection and not for 3D reconstruction, we design network structure and complexity to be a good compromise between accuracy and execution speed.
\\
\subsection{Depth pixel-wise estimation as an encoder-decoder network}
Our proposed architecture is reminiscent of fully convolutional architectures as \cite{fischer2015flownet}, \cite{badrinarayanan2015segnet} and \cite{noh2015learning}. The encoder section is composed by a stack of convolutional layers, which apply learned filters on their input and extract relevant synthetic features. We do not apply naive pooling with an a-priori chosen strategy, such as max or average pooling. We choose instead to conveniently stride convolutions  in order to obtain a downsampled version of its input. Convolution output's dimensions $h_{conv}$ and $w_{conv}$, are determined, given an input of size $h \times w$, defining $k$ as the convolution kernel size, $p$ as convolution pad and $s$ as applied stride, by the following equations:

\begin{equation}
	h_{conv} = \frac{h + 2*p_{h}-k_{h}}{s}+1
\end{equation} 
\begin{equation}
	w_{conv} = \frac{w + 2*p_{w}-k_{w}}{s}+1
\end{equation} 
From these equations we can infer how, choosing appropriate stride and padding, we are able to downsample information directly from convolutions, allowing the network to learn the optimal scaling strategy according to the task at hand. 
The decoder section is composed by a stack of deconvolutional layers, which learns to upsample from the features computed in the encoder section to obtain a final output of the same resolution of the input, containing pixel-wise predictions. Other works as \cite{badrinarayanan2015segnet} or \cite{noh2015learning} place unpooling layers between each deconvolutional layer to reverse pooling operations done in the encoder section; since downsampling is performed by convolution layers, we model our deconvolutional layers and learn the most effective upsampling strategy, as an inverse operation.
\\ 
Detailed network implementation is shown on Image \ref{fig:network}. Encoder section is composed by five $3\times 3$ convolutional layers. Strided convolutions are applied at the first, second, third and fifth layer of the encoder section to downsample feature maps. Padding is added accordingly to maintain desired feature maps size. ReLU non-linearity is applied after each convolution output. At the end of the encoder section, we obtain feature maps downscaled by a factor of 16 compared to network input. \\
Decoder section is composed by three deconvolutional layer. Each deconvolutional layer learn to upsample encoder feature maps by, respectively, a factor of 2 for the first two layers and a factor of 4 for the final layer, in order to obtain a final upsampling factor of 16. In \cite{fischer2015flownet} and \cite{long2015fully} feature maps computed in intermediate convolutional layers in the encoder section are concatenated to each intermediate deconvolutional layer output to improve upsampling quality and edges definitions. We experimented this strategy in preliminary experiments; although we came upon a slight improvement on upsampling quality, we also experimented a performance degradation in terms of inference time, so we did not apply this strategy in successive experiments.
\\
\subsection{Image and optical flow as network input} \label{sec:inputs}
In order to choose appropriate network input, we compare in our experiments two possible strategies: feeding the network with a single image, currently captured by the camera, or concatenate current image with optical flow information between current frame and the previous one. Optical flow has been used previously as raw feature for obstacle detectors \cite{dey2014vision} \cite{beyeler2009vision}. It is known how relative motion information between each pixel in two consecutive frames contains some implicit information about object dimensions and locations in 3D space. As previous works stated, optical flow alone is not sufficient to obtain a complete and long-range depth estimation. Our intuition is to use it as additional information and let convolutional filters learn optimal strategy to extract useful information from it. Mixing together optical flow and raw image as network input, we expect them to overcome each other's limitations and improve performances and generalization capability in real world scenarios.
Optical flow during our experiments has been computed off-line using widely-used  and robust Brox algorithm \cite{Brox2004}, but faster and effective algorithms as \cite{fischer2015flownet} may be used as well to improve whole software pipeline real-time performance.

\begin{figure*}[ht!]
\centering
  \ffigbox{}
  {
    \CommonHeightRow
    {
      \begin{subfloatrow}[3]
      \hspace{-0.8em}
    \ffigbox[\FBwidth]
    {\includegraphics[height=\CommonHeight]{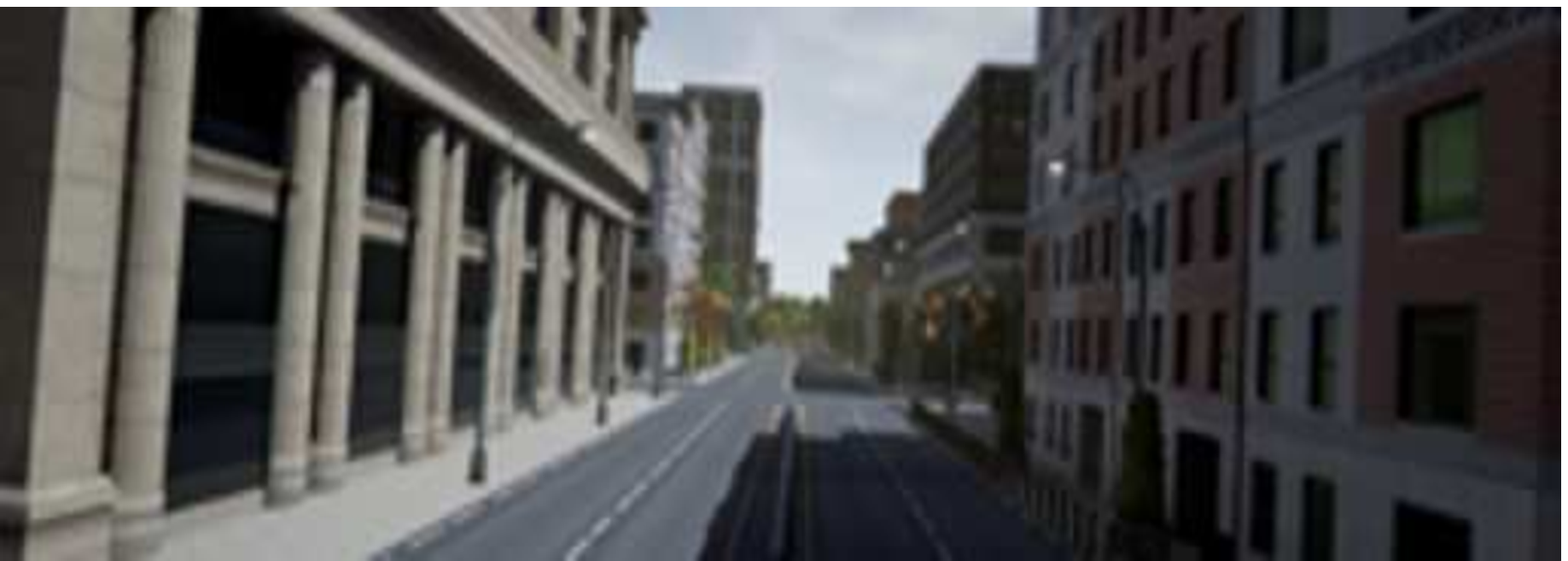}}
    {\vspace{1.0 em}}
    \ffigbox[\FBwidth]
    {\includegraphics[height=\CommonHeight]{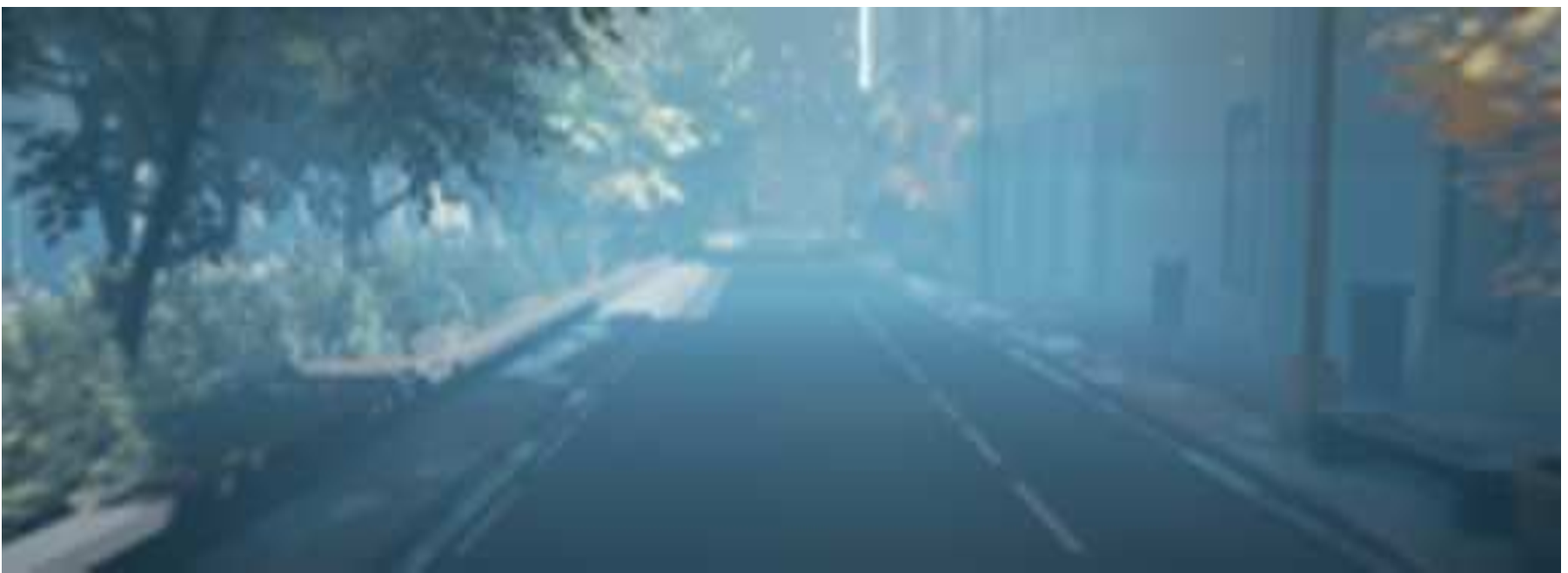}}
    {\vspace{1.0em}}
    \ffigbox[\FBwidth]
    {\includegraphics[height=\CommonHeight]{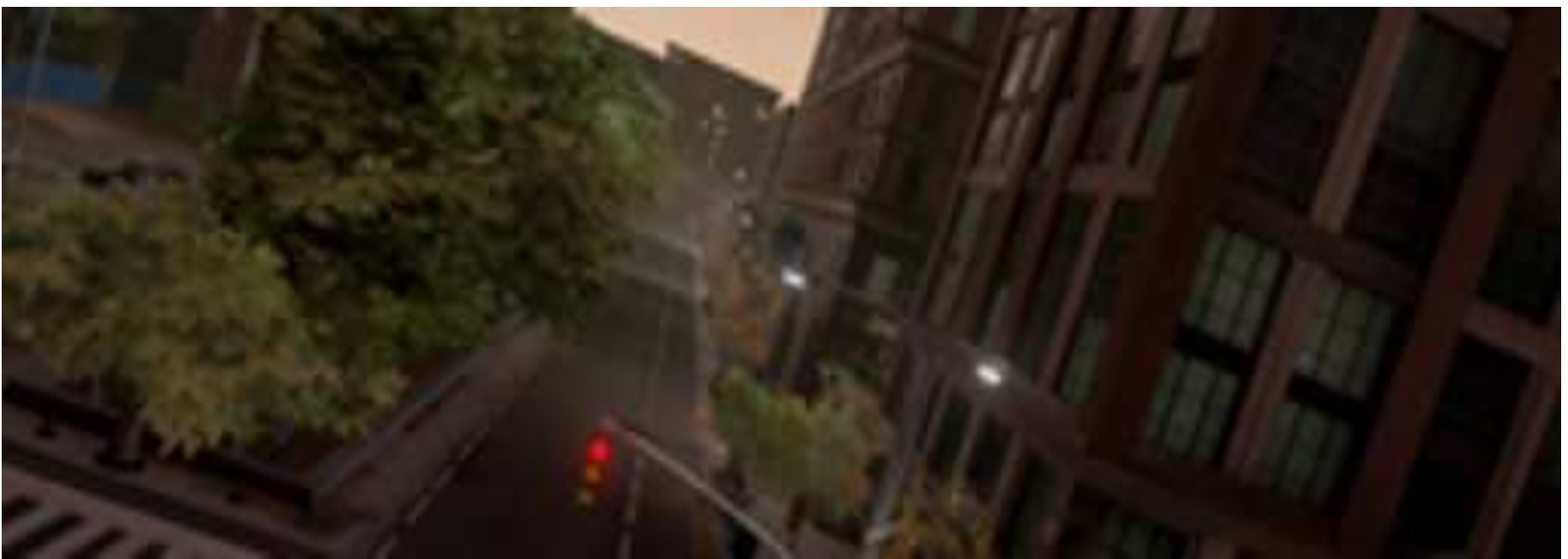}}
    {\vspace{1.0em}}

      \end{subfloatrow}
    }    
    \CommonHeightRow
    {
      \begin{subfloatrow}[3]
      \hspace{-0.8em}
    \ffigbox[\FBwidth]
    {\includegraphics[height=\CommonHeight]{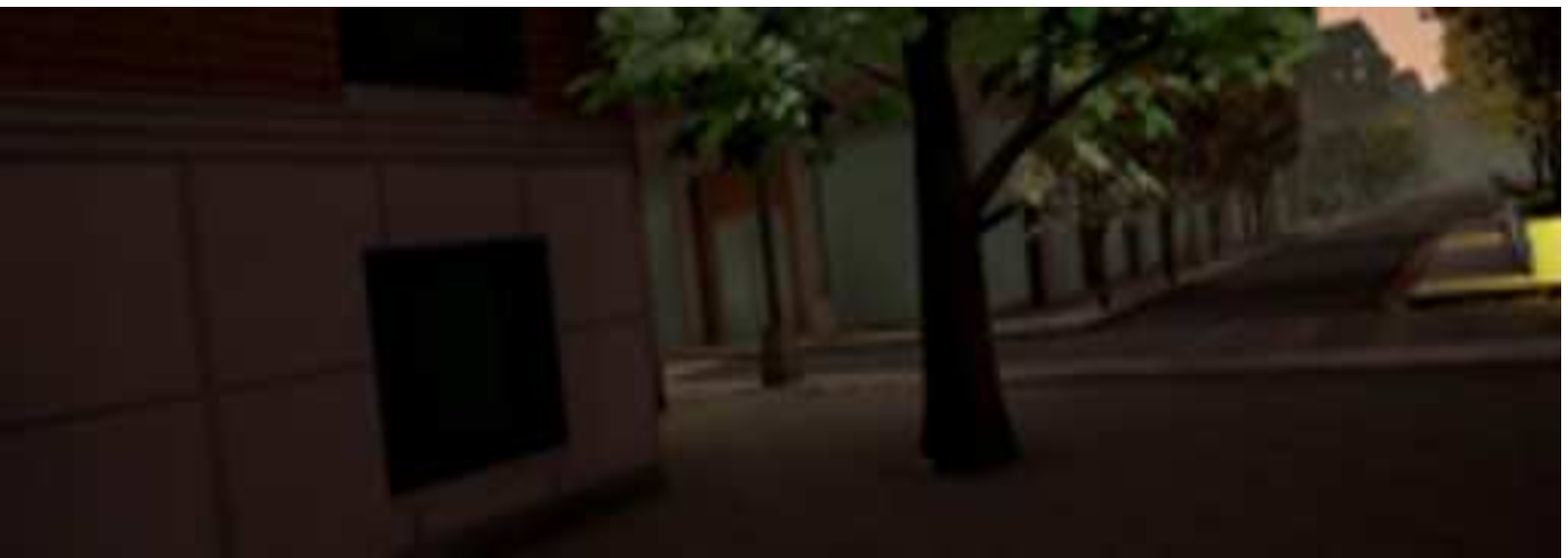}}
    {\vspace{1.0em}}
    \ffigbox[\FBwidth]
    {\includegraphics[height=\CommonHeight]{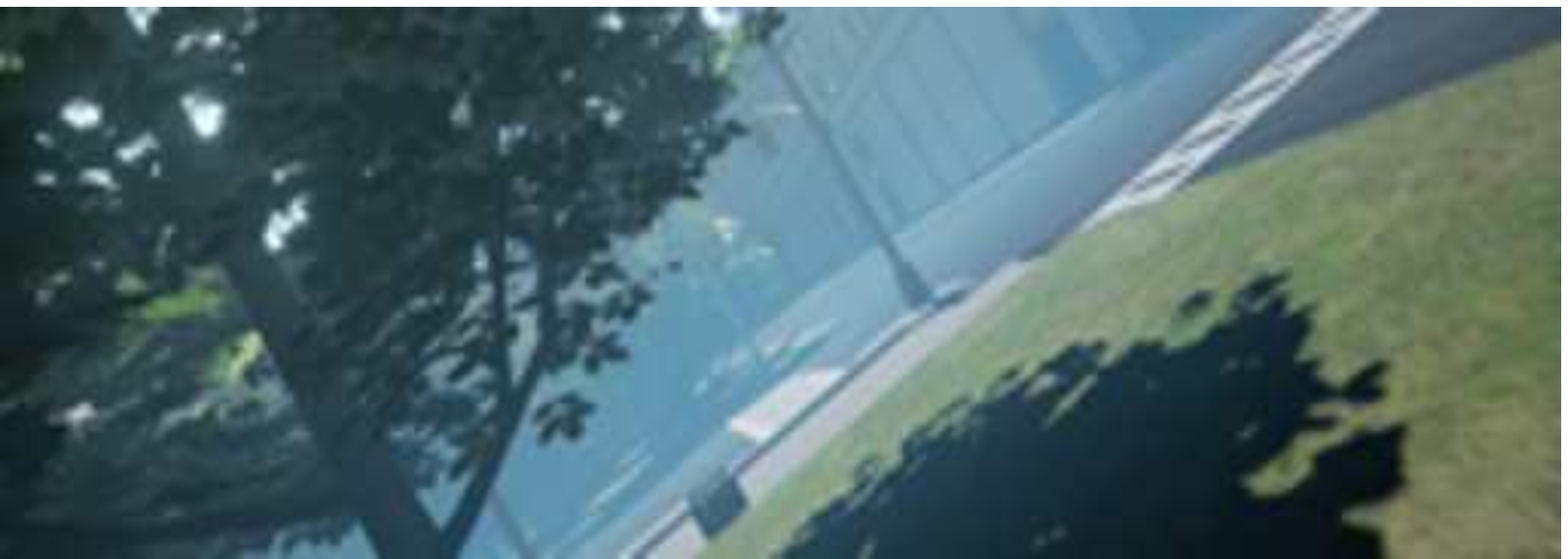}}
    {\vspace{1.0em}}
    \ffigbox[\FBwidth]
    {\includegraphics[height=\CommonHeight]{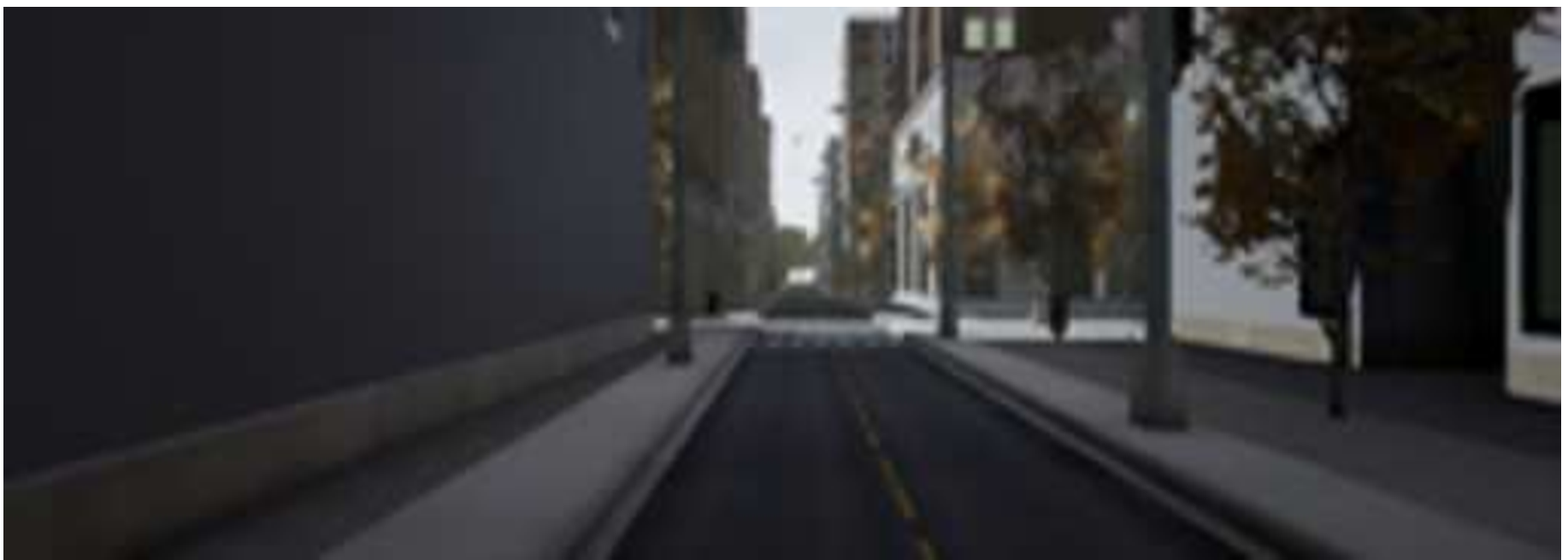}}
    {\vspace{1.0em}}

      \end{subfloatrow}
    } 
    

    \vspace{-2em}
      \caption{\small Some images from Virtual Dataset, highlighting lighting 		conditions and captured motion's diversity comprised into the dataset \vspace{-1.3em}}  
       \label{fig:vd_images}
    
    }
\end{figure*}
\subsection{Virtual Dataset} \label{sec:virtual_dataset}

Collection of a sufficient amount of training data is a typical problem for every deep learning work.
Considering how we formulate our problem of finding obstacles, we explore existing datasets containing depth ground truth. NYUDepth v2 indoor dataset  \cite{silberman2012indoor}, Make3D \cite{saxena2009make3d} and KITTI outdoor datasets \cite{Geiger12} are typical choices for depth estimation-related problems. Make3D and NYUDepth datasets contain still images of a scene with no sequentiality between them, since they are thought for 3D image reconstruction, and this makes our optical flow-based approach not applicable.
KITTI sequences are grabbed by a camera mounted on a moving car, making it more appropriate for robotic applications. KITTI depth ground truth, collected through a LiDAR unit, is sparse and does not cover the whole image scene. Moreover, images are generally aligned with ground plane, which may be a limitation according to the desired operating scenarios. \\
Motivated by these reasons, we explored the possibility to collect data from virtual scenarios, utilizing development tools generally used in gaming industry, exploiting capabilities of the newest graphic engines.  
We utilize Unreal Engine 4 with Urban City pack developed by PolyPixel to build an urban scenario sized about $0.36$ km$^{2}$. No car, person, or dynamic object is present due to development package limitations; their inclusion will be considered as future work. We move a camera, collecting images and dense scene depth ground truth. Camera moves around the virtual world with six degrees of freedom, simulating non-trivial movements that rarely are present in real-world datasets, such as huge roll or pitch angles with respect to the ground plane. 
Depth is stored as a grayscale image: it is converted into metric depth by scaling each pixel value by a factor obtained through placing objects in a toy scenario at known metric distance from the camera. For preliminary experiments, depth is collected firstly with a maximum range of 40 meters; in a second phase, we collected depth measures up to 200 meters. Depth measures are spherical with respect of the camera. 
\\
More that 265k images have been collected and stored in PNG format, with a resolution of $1241 \times 376 $ pixels. Light conditions are changed from time to time to achieve brightness robustness. Haze is added in some sequences as well, and motion blur is simulated through graphic engine's tools, in order to better simulate real scenarios. Images are collected as sequences of consecutive frames captured as the camera moves around the world, at a rate of 10 Hz. We move the camera both on-road and off-road environments, for example between trees or light poles, to better simulate possible realistic application scenarios.

\section{Experiments}\label{sec:exp}

To validate our work, we perform experiments on our Virtual Dataset, as described on Section \ref{sec:virtual_dataset}, as well as on KITTI dataset \cite{Geiger12}. 
We also test our network's estimation robustness adding artificial blurring and darkening on KITTI images, to evaluate network's performance in presence of noise.
For all of our experiments, we train the proposed networks on our Virtual Dataset, divided into a training set composed by about 200k images and a test set of 
65k images. In order to evaluate the generalization capabilities of our approach, we do not perform any fine-tuning on the KITTI sequences.
Training and testing are performed on a NVIDIA K40 GPU-mounted workstation. 

We update weights during training by using Stochastic Gradient Descent (SGD) algorithm with a learning rate $\alpha = 10^{-3}$, gradually scaled down during training. 
Convergence is reached after about 50 epochs on training data. We train our final proposed architecture on Log RMSE (\ref{eq:log_rmse}), in order to penalize more errors on close obstacles than ones committed on long range estimations: 

\begin{equation} \label{eq:log_rmse}
	\sqrt{\frac{1}{T}\sum_{Y\in T}||\log y_{i}-\log y_{i}^{*}||^{2}}  
\end{equation} 
Exploratory experiments are also performed with a linear RMSE loss, as specified later in Section \ref{sec:vd_exps}.
\\
Network inference time, without taking into account optical flow computation, is about 34 ms ($\mathbf{\sim300Hz}$) on K40 for each frame, which allows optimal scalability into complete embedded robotic software pipelines. Brox's optical flow algorithm used for our experiments runs at about $10Hz$, but much faster algorithms, as \cite{fischer2015flownet}, could be used as well.

The benchmark metrics for our comparisons are:
\begin{itemize}
\item Threshold error: \% of $y_{i}$ s.t. $\max (\frac{y_{i}}{y_{i}^{*}}\frac{y_{i}^{*}}{y_{i}}) = \delta < thr$ 
\\
\item Linear RMSE: $\sqrt{\frac{1}{T}\sum_{Y\in T}||y_{i}-y_{i}^{*}||^{2}}  $
\\
\item Scale-invariant Log MSE (as introduced by \cite{eigen2014depth}):
$ \frac{1}{n}\sum_{i} d_{i}^{2} - \frac{1}{n^{2}}(\sum_{i} d_{i})^{2}$, with $d_{i} = \log y_{i}-\log y_{i}^{*}$
\end{itemize}

\subsection{Virtual Dataset} \label{sec:vd_exps}

\begin{table}[hb!]
  \centering
  \caption{\small Experiments results on virtual dataset for ground truth collected up to 40m.}
  \label{tab:40_meters_vd_results}
  \begin{tabular}{|c||c|c|c|}
    \hline
     & \small Single Image & \small Opt. Flow+Img.& \\
    \hline
    \small thr. $\delta<1.25$ 		& 0.726 	& \textbf{0.774} & \small Higher\\
    \small thr. $\delta<1.25^{2}$ 	& 0.924 	& \textbf{0.938} & \small is better\\
    \hline
    \small RMSE 					& 3.819		& \textbf{3.478} & \small Lower\\
    \small Log RMSE 				& 0.246		& \textbf{0.221} & \small is\\
    \small Scale Inv. MSE 			& 0.065		& \textbf{0.055} & \small better\\
    \hline
  \end{tabular}
\end{table}

We initially perform exploratory experiments on Virtual Dataset test set to compare the performance of the two proposed architectures (as described in Section \ref{sec:inputs}).
Networks are trained by using sequences with ground truth depth collected up to 40 meters, using Linear RMSE as training loss. 
Quantitative results are shown in Table \ref{tab:40_meters_vd_results}: the network that processes optical flow inputs outperforms single image network with respect to all the metrics, showing the
effectiveness of the proposed optical flow+image network.

Afterwards, we compute metric depth up to 200 meters and re-train the optical flow-based architecture on log RMSE and linear RMSE (see Table \ref{tab:log_vs_linear}). 
The results show that the network based on the log RMSE loss achieve better performance with respect to the linear one.

\begin{table}[h!]
  \centering
  \caption{\small Comparison between the optical flow+image network trained on Log RMSE and the linear RMSE}
  \label{tab:log_vs_linear}
  \begin{tabular}{|c||c|c|c|}
    \hline
     						& Log RMSE & Linear RMSE & \\
    \hline
    \small thr. $\delta<1.25$ 		&  \textbf{0.643}	& 0.482 & \small Higher\\
    \small thr. $\delta<1.25^{2}$ 	&  \textbf{0.887}	& 0.764 & \small is better\\ \hline
    \small RMSE 					&  \textbf{6.065}	& 7.004 & \small Lower\\
    \small Log RMSE 				&  \textbf{0.292}	& 0.416 & \small is\\
    \small Scale Inv. MSE 			&  \textbf{0.085}	& 0.154 & \small better\\ 
    \hline
  \end{tabular}
\end{table}

\begin{figure*}[ht!]
\centering
  \ffigbox{}
  {
    \CommonHeightRow
    {
      \begin{subfloatrow}[4]
      \hspace{-0.8em}
    \ffigbox[\FBwidth]
    {\includegraphics[height=\CommonHeight]{images/vd_img7.eps}}
    {\vspace{1em}}
    \ffigbox[\FBwidth]
    {\includegraphics[height=\CommonHeight]{images/vd_img9.eps}}
    {\vspace{1em}}
    \ffigbox[\FBwidth]
    {\includegraphics[height=\CommonHeight]{images/vd_img5.eps}}
    {\vspace{1em}}
    
    \ffigbox[\FBwidth]
    {\includegraphics[height=\CommonHeight]{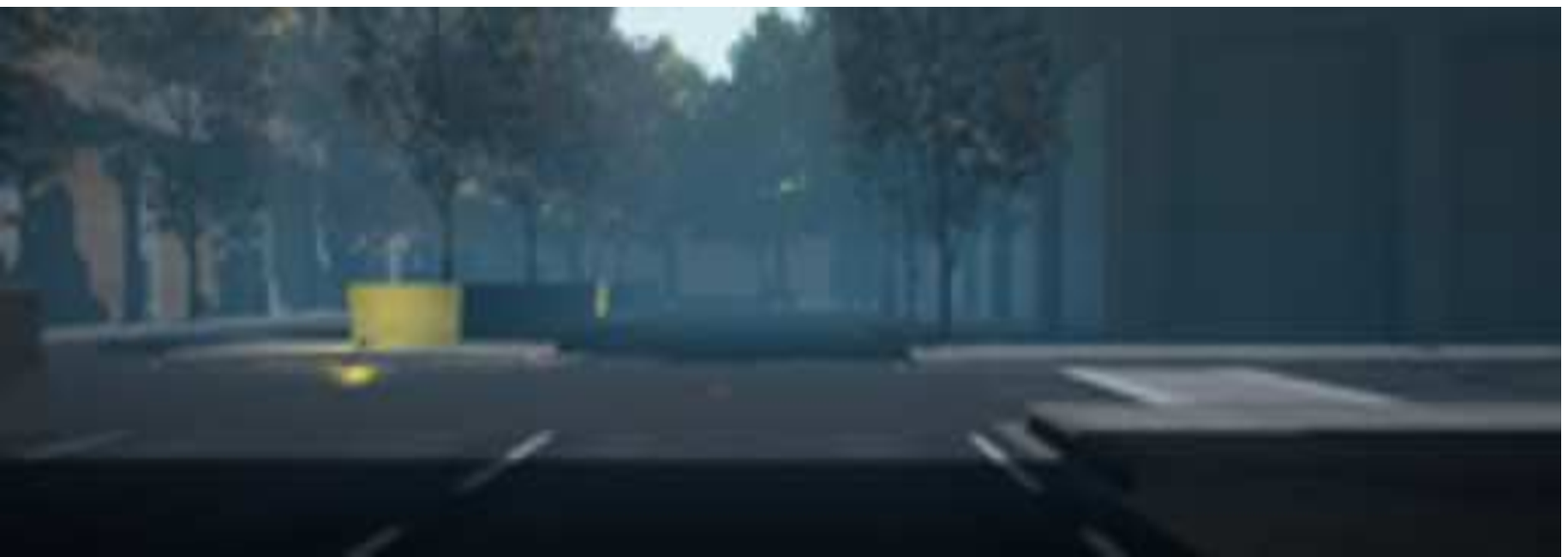}}
    {\vspace{1em}}

      \end{subfloatrow}
    }    
    \CommonHeightRow
    {
      \begin{subfloatrow}[4]
      \hspace{-0.8em}
    \ffigbox[\FBwidth]
    {\includegraphics[height=\CommonHeight]{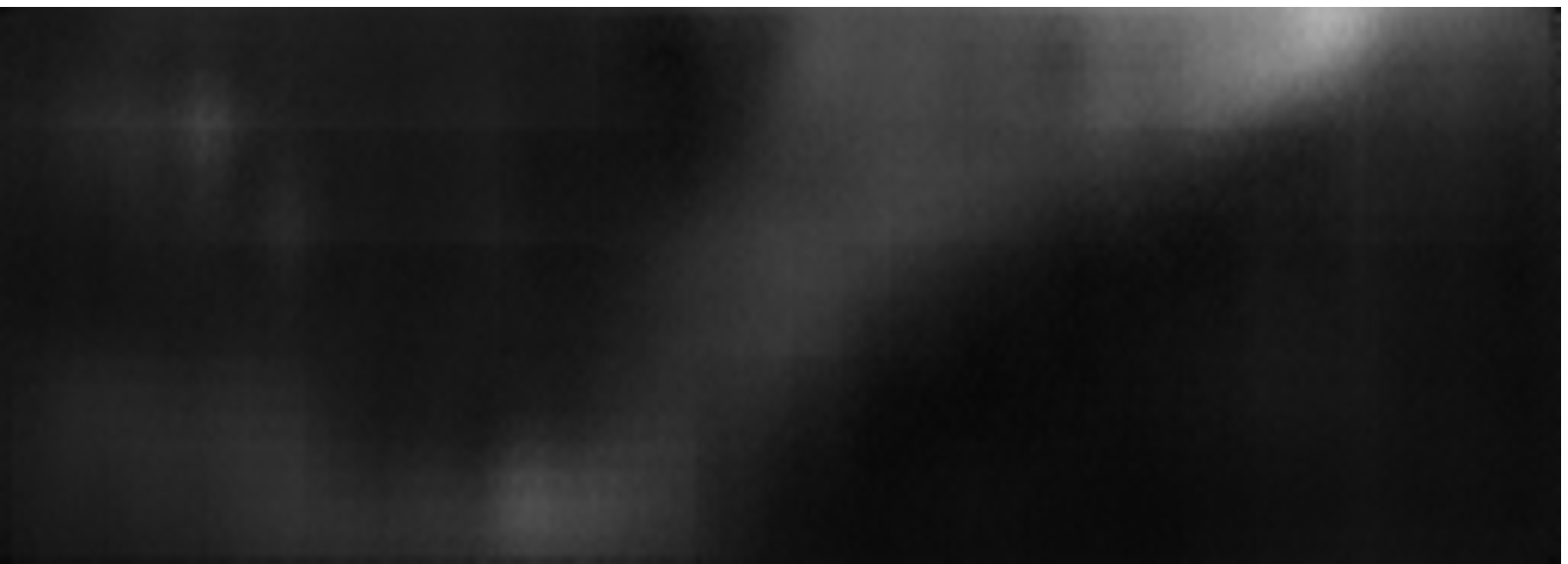}}
    {\vspace{1em}}
    \ffigbox[\FBwidth]
    {\includegraphics[height=\CommonHeight]{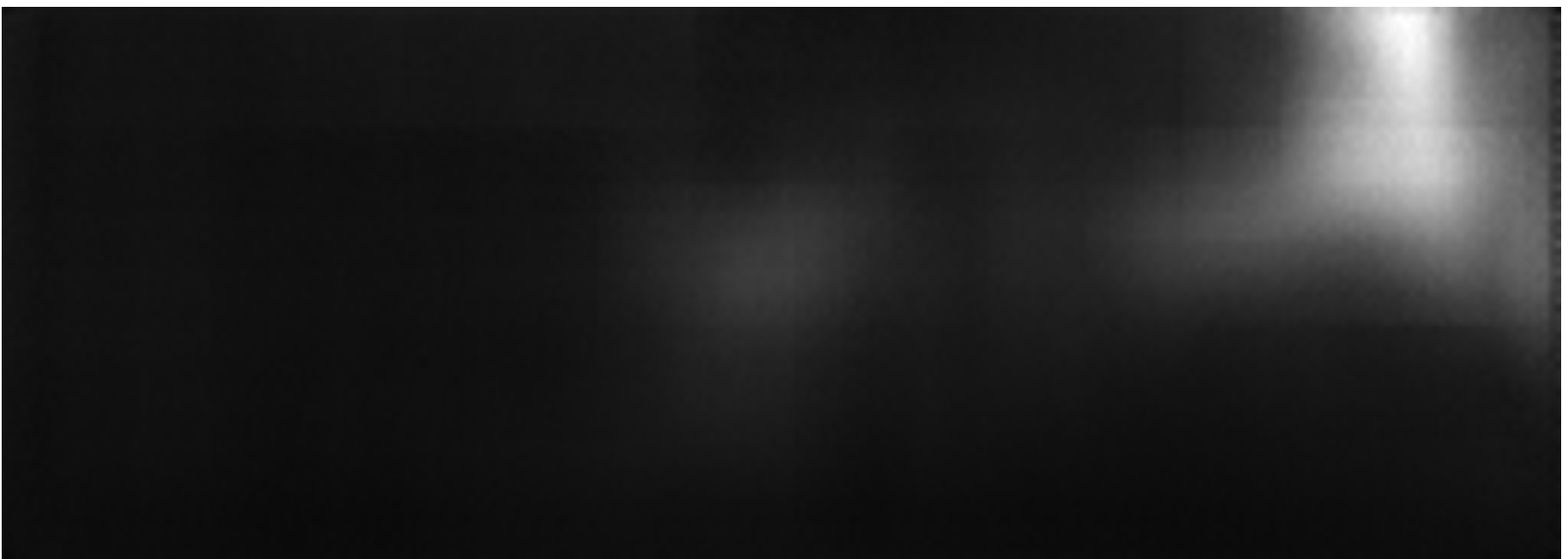}}
    {\vspace{1em}}
    \ffigbox[\FBwidth]
    {\includegraphics[height=\CommonHeight]{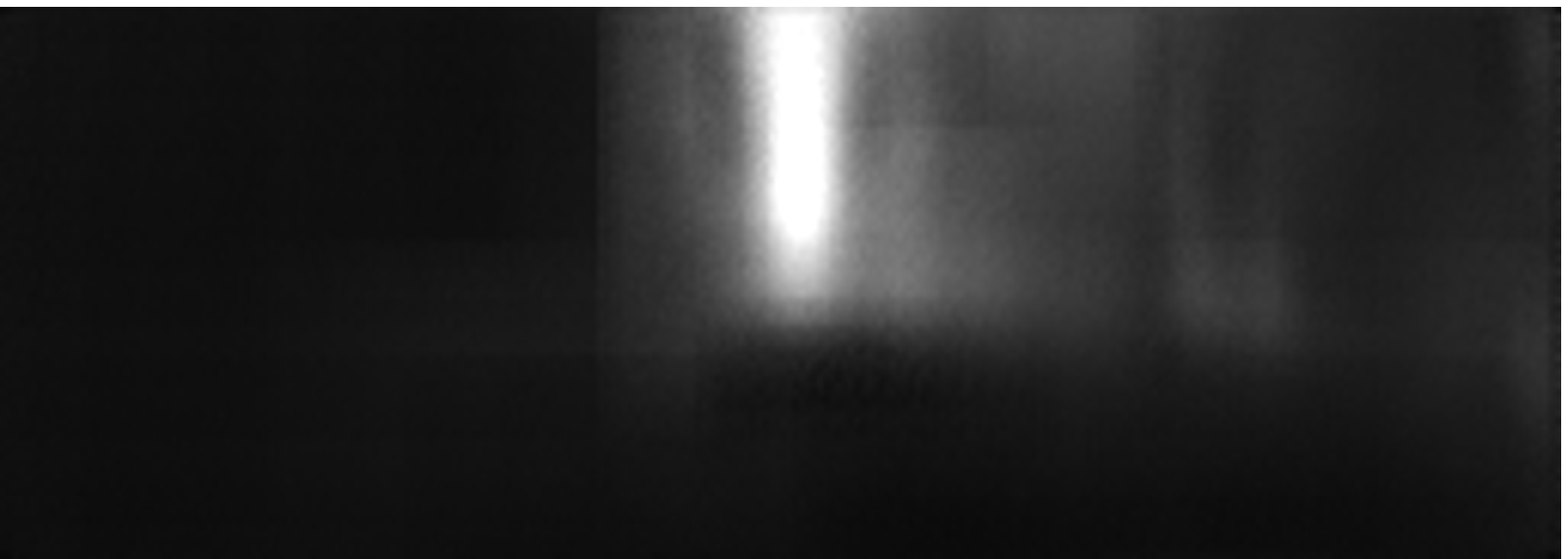}}
    {\vspace{1em}}
    \ffigbox[\FBwidth]
    {\includegraphics[height=\CommonHeight]{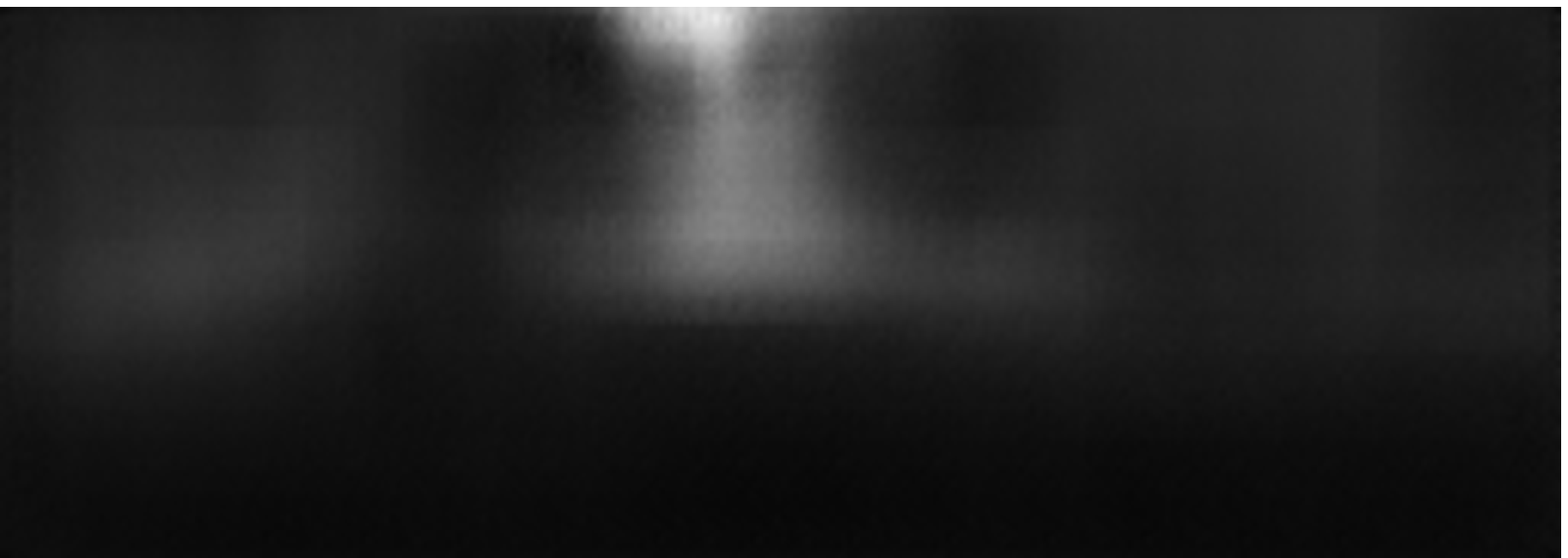}}
    {\vspace{1em}}

      \end{subfloatrow}
    } 
    \CommonHeightRow
    {
      \begin{subfloatrow}[4]
      \hspace{-0.8em}
    \ffigbox[\FBwidth]
    {\includegraphics[height=\CommonHeight]{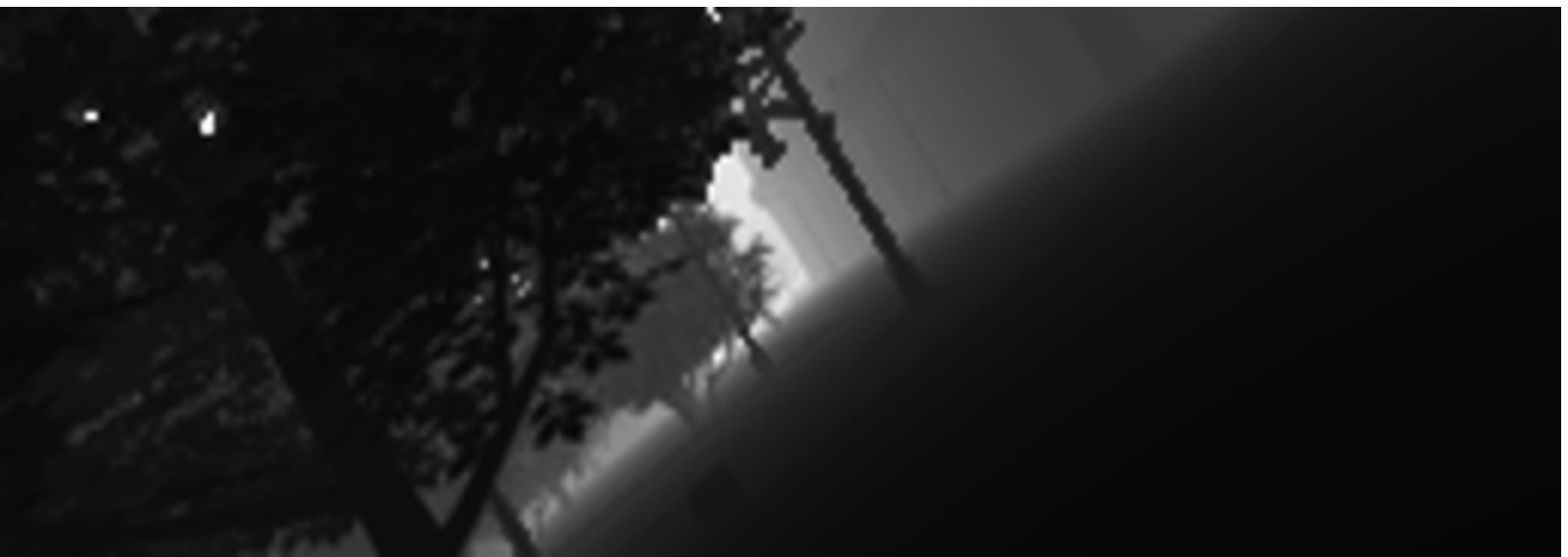}}
    {\vspace{1em}}
    \ffigbox[\FBwidth]
    {\includegraphics[height=\CommonHeight]{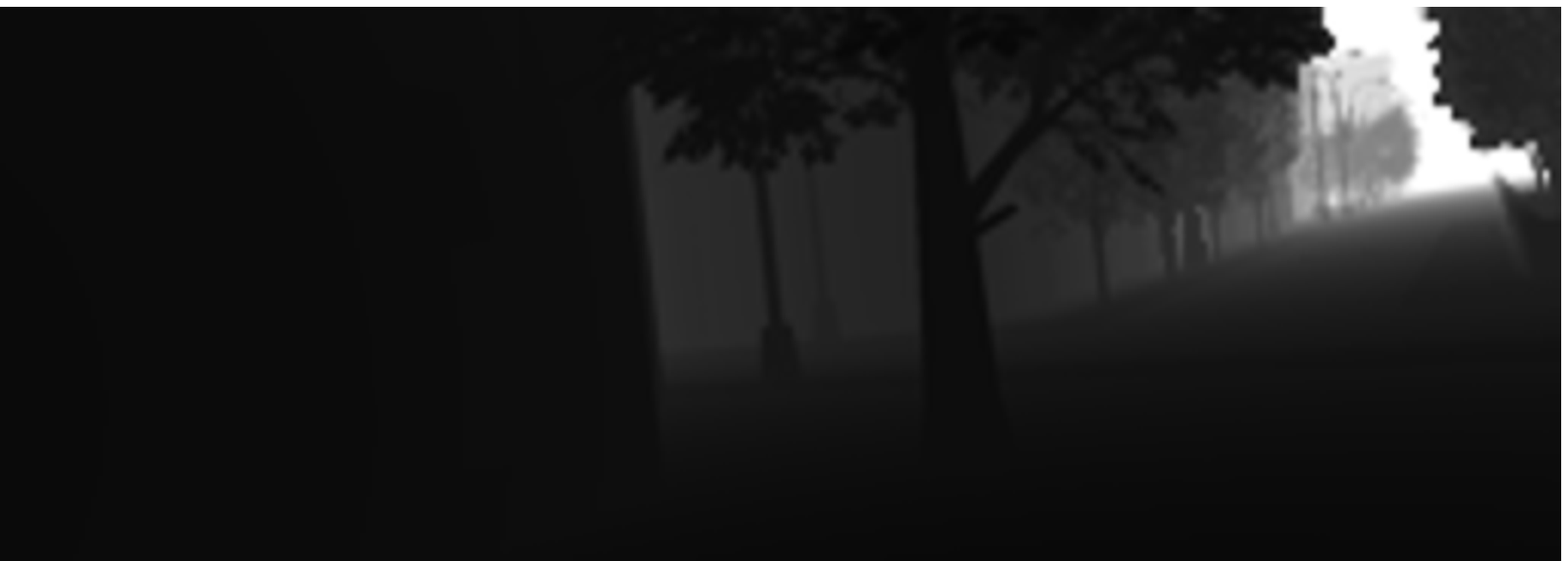}}
    {\vspace{1em}}
    \ffigbox[\FBwidth]
    {\includegraphics[height=\CommonHeight]{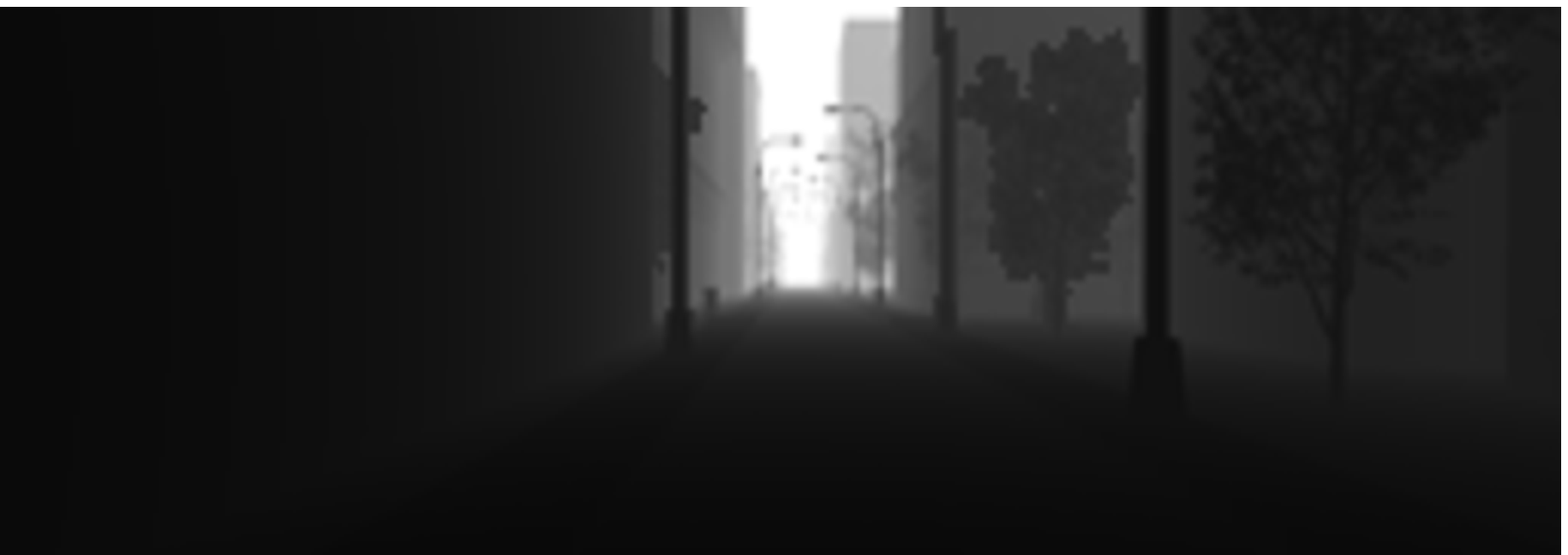}}
    {\vspace{1em}}
    \ffigbox[\FBwidth]
    {\includegraphics[height=\CommonHeight]{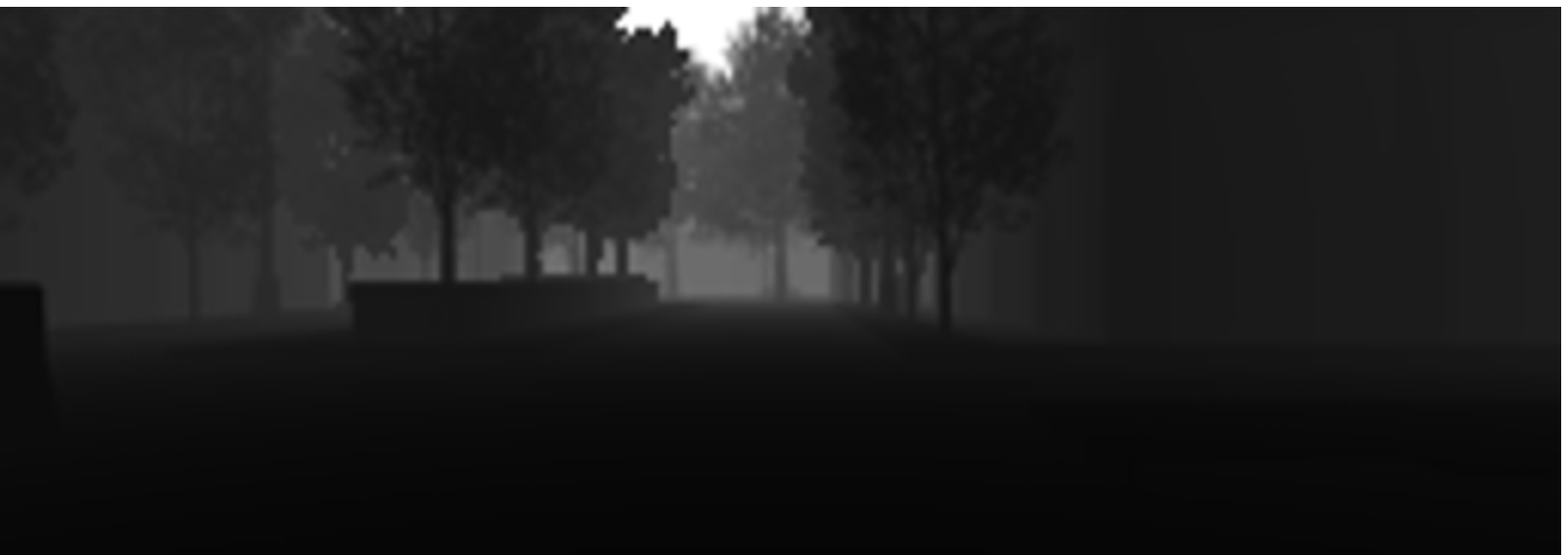}}
    {\vspace{1em}}

      \end{subfloatrow}
    } 
    \vspace{-1.5em}
      \caption{\small Qualitative results on the Virtual Dataset. On the first row RGB input images are depicted. The second and the third rows show the network predictions and the dense ground truths, respectively.\vspace{-1.3em}}  
       \label{fig:vd_results}
    }
 
\end{figure*}


\subsection{KITTI dataset}

We perform experiment on KITTI dataset \cite{Geiger12}. The sequences are gathered with a Pointgrey Flea2 firewire stereo camera mounted on a car traveling in the streets of the Karlsruhe city.
Images are undistorted and collected with a resolution of $1240\times386$ and a frame rate of 10Hz.
As the provided depth ground truth is sparse, we compute dense ground truth by using the colourization routine proposed in \cite{silberman2012indoor}. 
Furthermore, since LiDAR provides ground truth measures only for the bottom half of the scene, experiments are performed with respect to that portion of data. 
Performance are evaluated on a test set composed by 697 images, corresponding to the published results of \cite{eigen2014depth}.
We first run exploratory experiments to evaluate the optical flow based architecture, re-using network weights trained on a 
maximum detection range up to 40 meters, as described in Section \ref{sec:vd_exps}. Results are provided in Table \ref{tab:kitti_40}.

\begin{table}[h!]
  \centering
  \caption{\small Results on KITTI Dataset on our architectures trained with a detection range$ < 40$ meters}
  \label{tab:kitti_40}
  \begin{tabular}{|c||c|c|c|}
    \hline
     						& \small Single Img. & \small Optical Flow+Img. & \\
    \hline
    \small thr. $\delta<1.25$ 		& 0.311 	& \textbf{0.421} & \small Higher\\
    \small thr. $\delta<1.25^{2}$ 	& 0.572 	& \textbf{0.679}  & \small is\\
    \small thr. $\delta<1.25^{3}$ 	& 0.764 	& \textbf{0.813} & \small better\\
    \hline
    \small RMSE 					& 7.542	& \textbf{6.863} & \small Lower\\
    \small Log RMSE 				& 0.574	& \textbf{0.504} & \small is\\
    \small Scale Inv. MSE 			& 0.206	& \textbf{0.205} & \small better\\
    \hline
  \end{tabular}
\end{table}

\begin{table*}[ht!]
\vspace{1.2em}
  \centering
  \caption{\small Results on KITTI Dataset}
  \label{tab:kitti}
  \begin{tabular}{|c||c|c|c|c|}
    \hline
     						& Our network & Eigen et. al \cite{eigen2014depth} & Saxena et al. \cite{saxena2009make3d} & \\
    \hline
    \small thr. $\delta<1.25$ 		& 0.318 	& \textbf{0.692} & 0.601 & \small Higher\\
    \small thr. $\delta<1.25^{2}$ 	& 0.617 	& \textbf{0.899} & 0.820 & \small is\\
    \small thr. $\delta<1.25^{3}$ 	& 0.813 	& \textbf{0.967} & 0.926 & \small better\\
    \hline
    \small RMSE 					& 7.508 & \textbf{7.156} & 8.734 & \small Lower\\
    \small Log RMSE 				& 0.524	& \textbf{0.270} & 0.361 & \small is\\
    \small Scale Inv. MSE 			& \textbf{0.196}	& 0.246 & 0.327 & \small better\\
    \hline
  \end{tabular}
\end{table*}

\begin{figure*}[ht!]
\centering
  \ffigbox{}
  {
    \CommonHeightRow
    {
      \begin{subfloatrow}[3]
      \hspace{-0.8em}
    \ffigbox[\FBwidth]
    {\includegraphics[height=\CommonHeight]{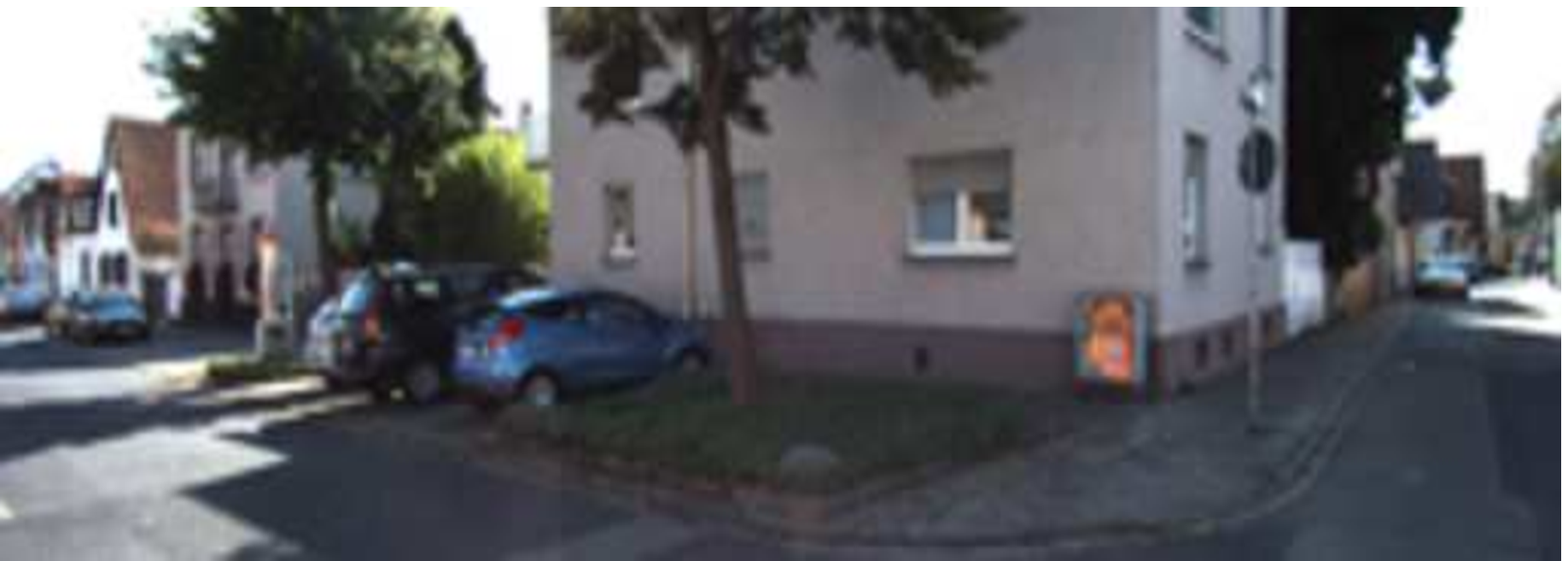}}
    {\vspace{1em}}
    \ffigbox[\FBwidth]
    {\includegraphics[height=\CommonHeight]{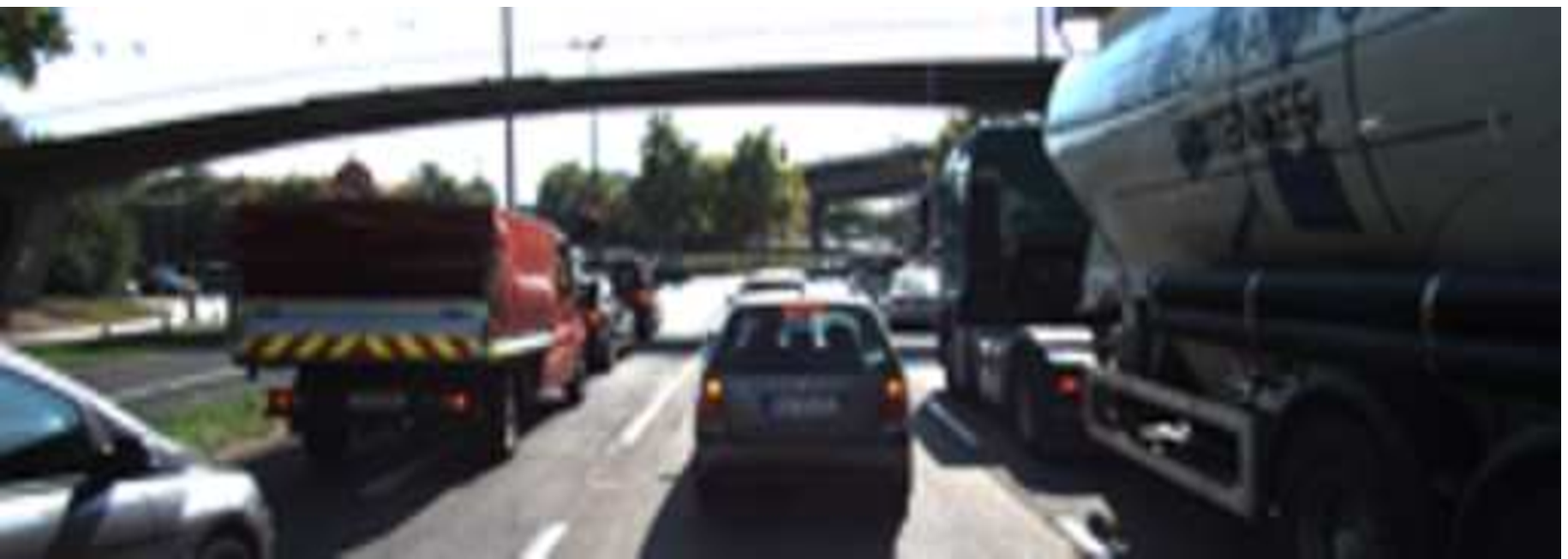}}
    {\vspace{1em}}
    \ffigbox[\FBwidth]
    {\includegraphics[height=\CommonHeight]{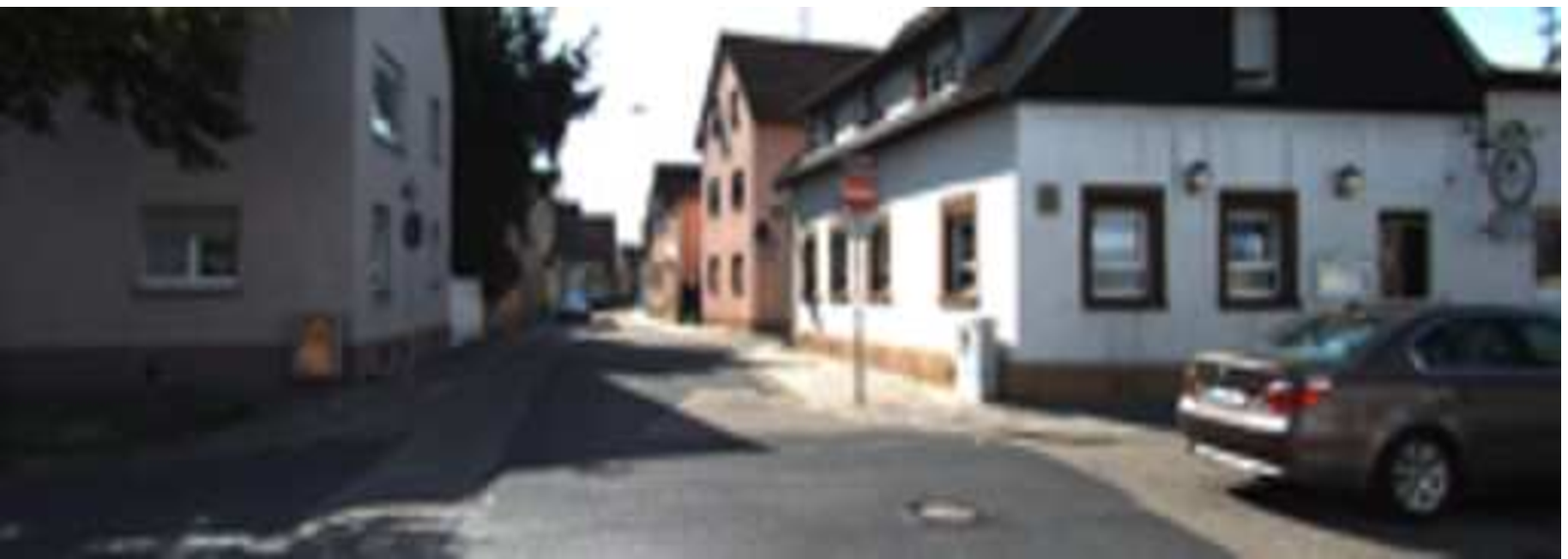}}
    {\vspace{1em}}

      \end{subfloatrow}
    }    
    \CommonHeightRow
    {
      \begin{subfloatrow}[3]
      \hspace{-0.8em}
    \ffigbox[\FBwidth]
    {\includegraphics[height=\CommonHeight]{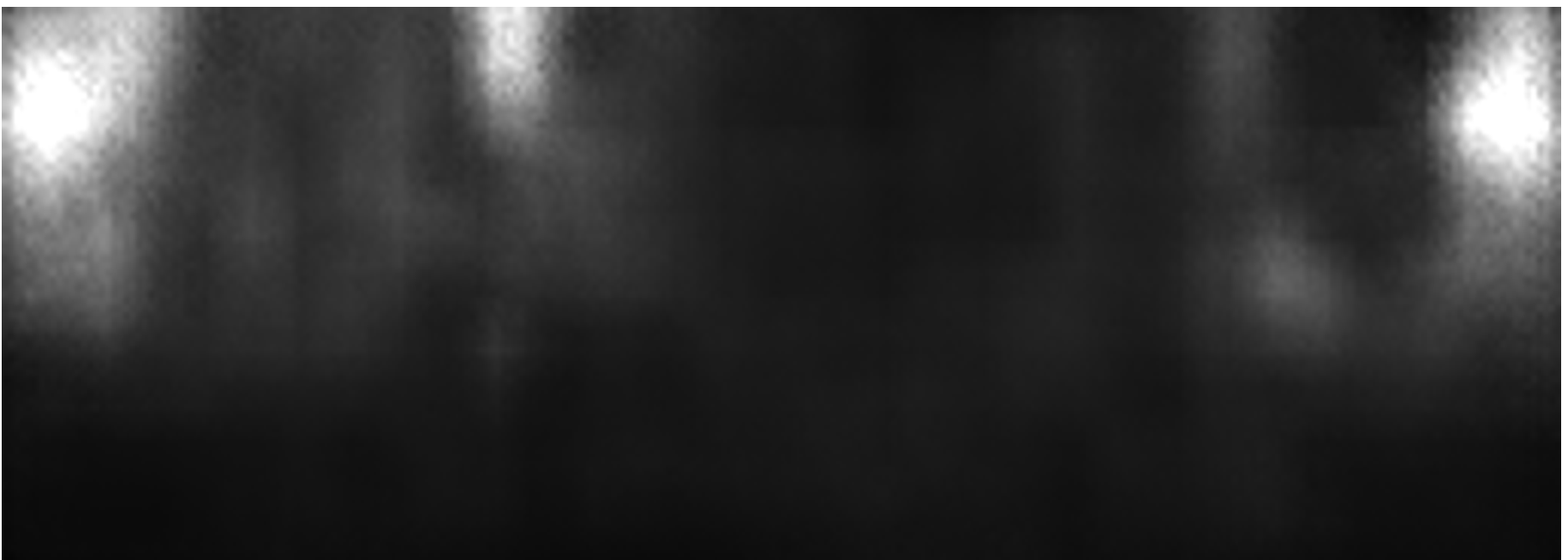}}
    {\vspace{1em}}
    \ffigbox[\FBwidth]
    {\includegraphics[height=\CommonHeight]{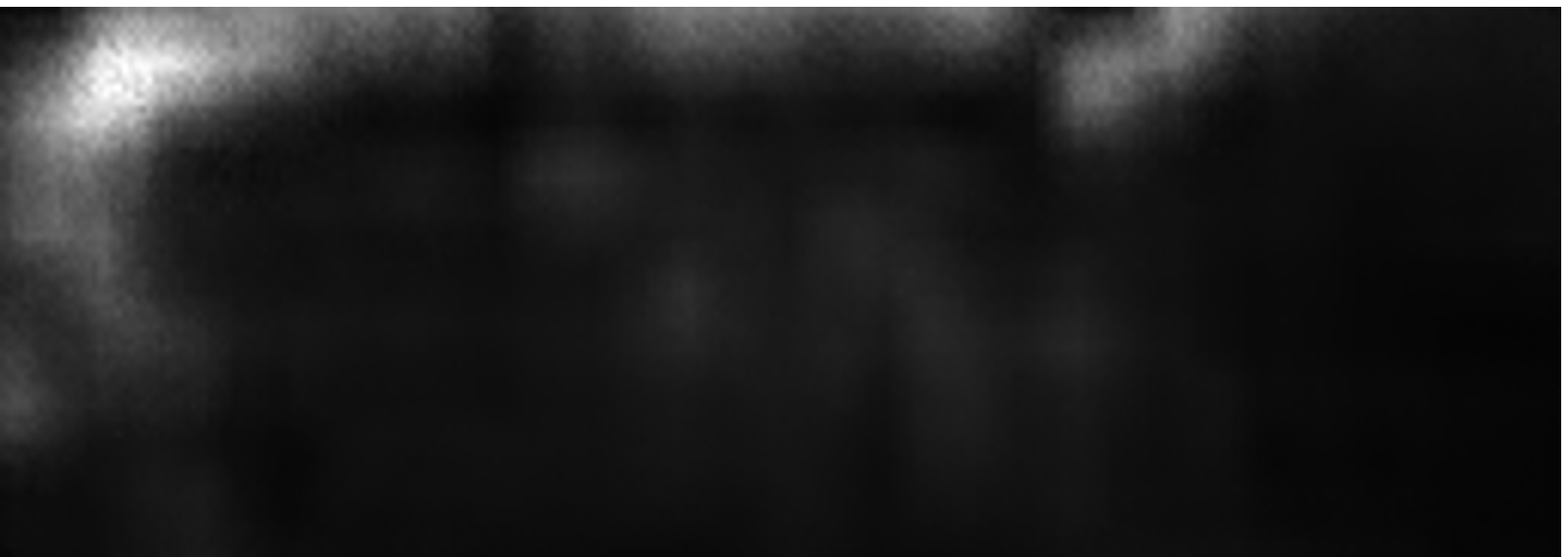}}
    {\vspace{1em}}
    \ffigbox[\FBwidth]
    {\includegraphics[height=\CommonHeight]{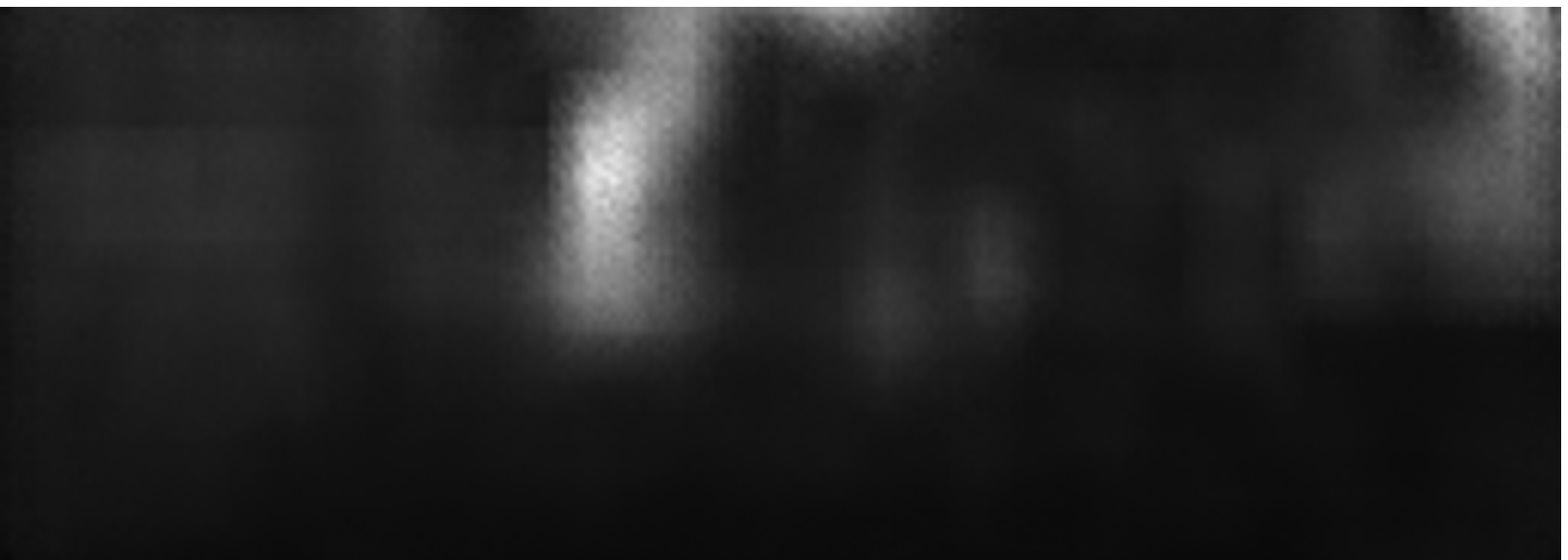}}
    {\vspace{1em}}

      \end{subfloatrow}
    } 
    \CommonHeightRow
    {
      \begin{subfloatrow}[3]
      \hspace{-0.8em}
    \ffigbox[\FBwidth]
    {\includegraphics[height=\CommonHeight]{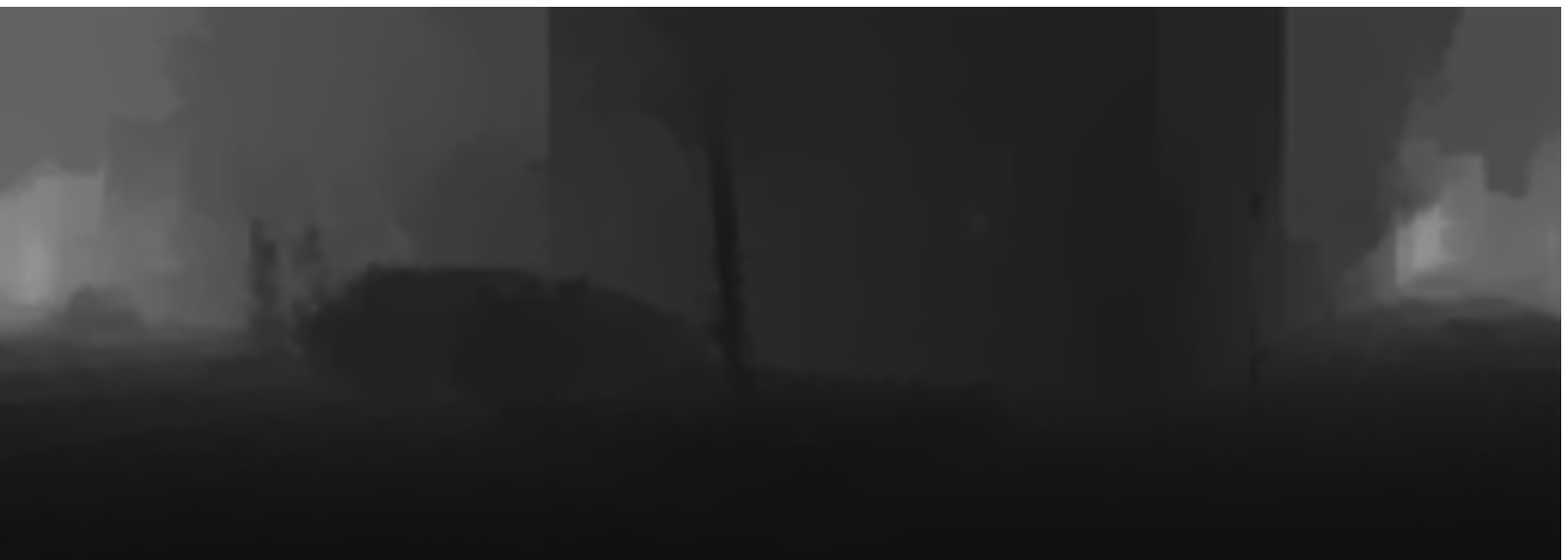}}
    {\vspace{1em}}
    \ffigbox[\FBwidth]
    {\includegraphics[height=\CommonHeight]{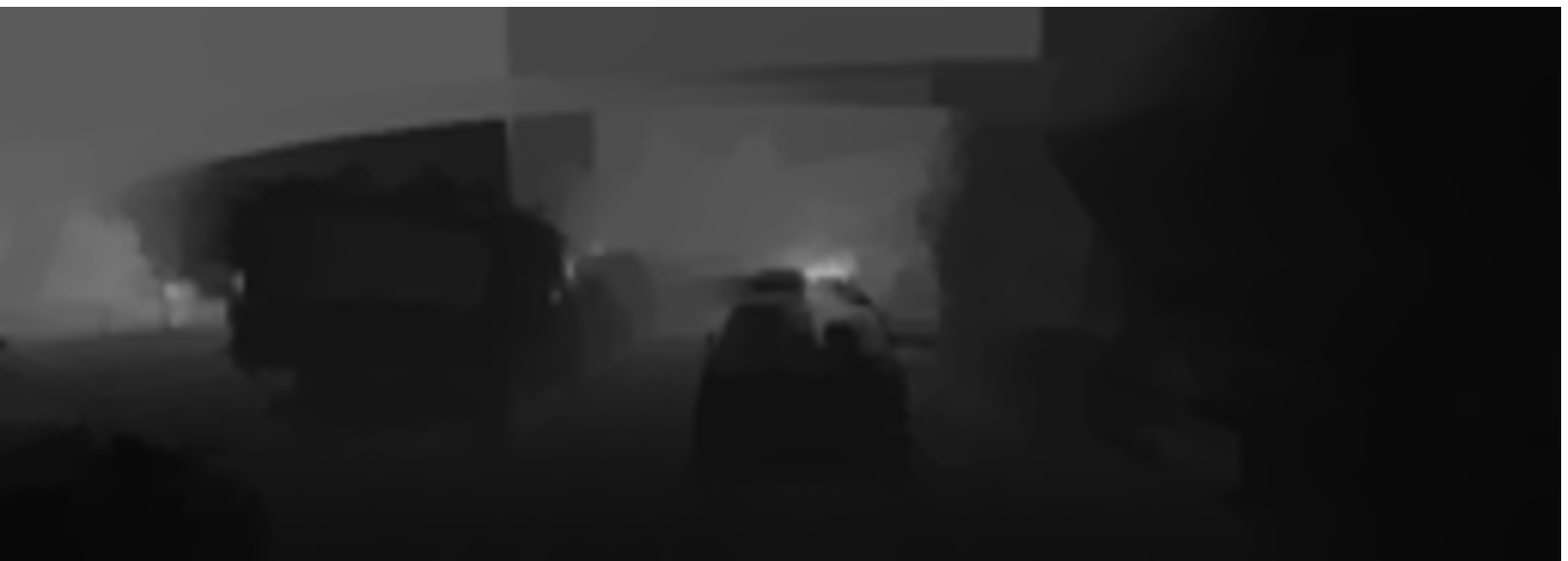}}
    {\vspace{1em}}
    \ffigbox[\FBwidth]
    {\includegraphics[height=\CommonHeight]{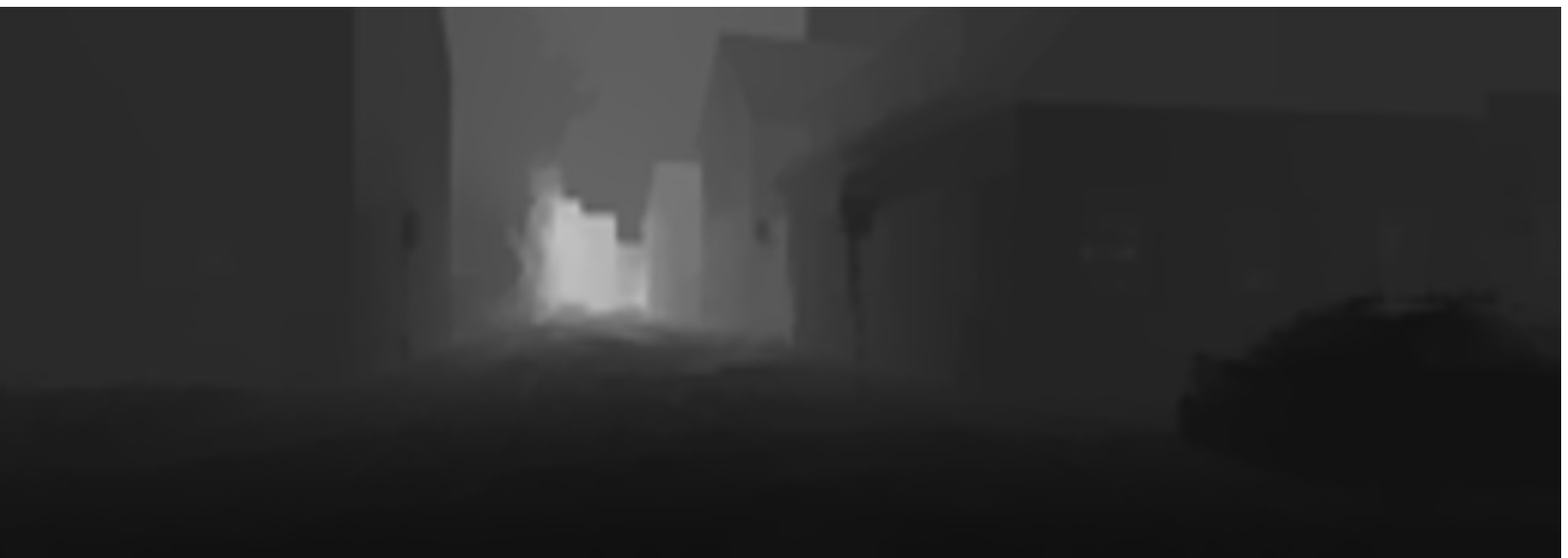}}
    {\vspace{1em}}

      \end{subfloatrow}
    } 
    \vspace{-1.5em}
      \caption{\small Qualitative results on KITTI dataset. The first row shows the input RGB image, while the second row and the third rows show the network prediction the dense ground truth obtained by using 
      the colourization routine, respectively. \vspace{-0.8em}}  
       \label{fig:kitti_results}    
    } 
\end{figure*}


Finally, we perform experiments on our optical flow-based network trained for detection up to 200 meters, and we compare its performance with respect to state of the art depth predictors. 
In particular, we compare our performances with Eigen et. al \cite{eigen2014depth} and Saxena et al. \cite{saxena2009make3d}.
It is important to notice that these approaches are both trained and tested with respect to the KITTI sequences. Conversely, in order to prove the generalization capabilities of our approach, 
we train our network with respect to the Virtual Dataset sequence and test on the KITTI sequences without any fine-tuning procedure.

We report results in Table \ref{tab:kitti}. For Saxena et al. work, we refer to the results provided in \cite{eigen2014depth}.
Although we do not perform any fine-tuning with respect to the real sequences, we obtain similar performance with respect to state-of-the-art approaches that are trained and tested on the same scenario.
Furthermore, our network outperforms the other methods with respect to the scale invariant Log MSE metric that penalizes relative scale errors without considering absolute scale imperfections. 
Hence, we infer that our approach provides a accurate estimates with respect to relative depths.

The performance that we achieved with respect to linear RMSE and log RMSE metrics, suggest that our network weaknesses lie on close-range estimations, 
as log RMSE penalizes more mistakes on small values. We acknowledge that, for obstacle detection tasks, there are more robust methods than can 
detect close obstacles, such as laser sensors or stereo cameras. Thus, we believe that our approach could definitely improve depth estimation performance
if combined with approaches tuned for short range detections.

\vspace{2em}

\begin{figure*}[ht!]
  \ffigbox{}
  {
    \CommonHeightRow
    {
      \begin{subfloatrow}[3]
      \hspace{-0.8em}
    \ffigbox[\FBwidth]
    {\includegraphics[height=\CommonHeight]{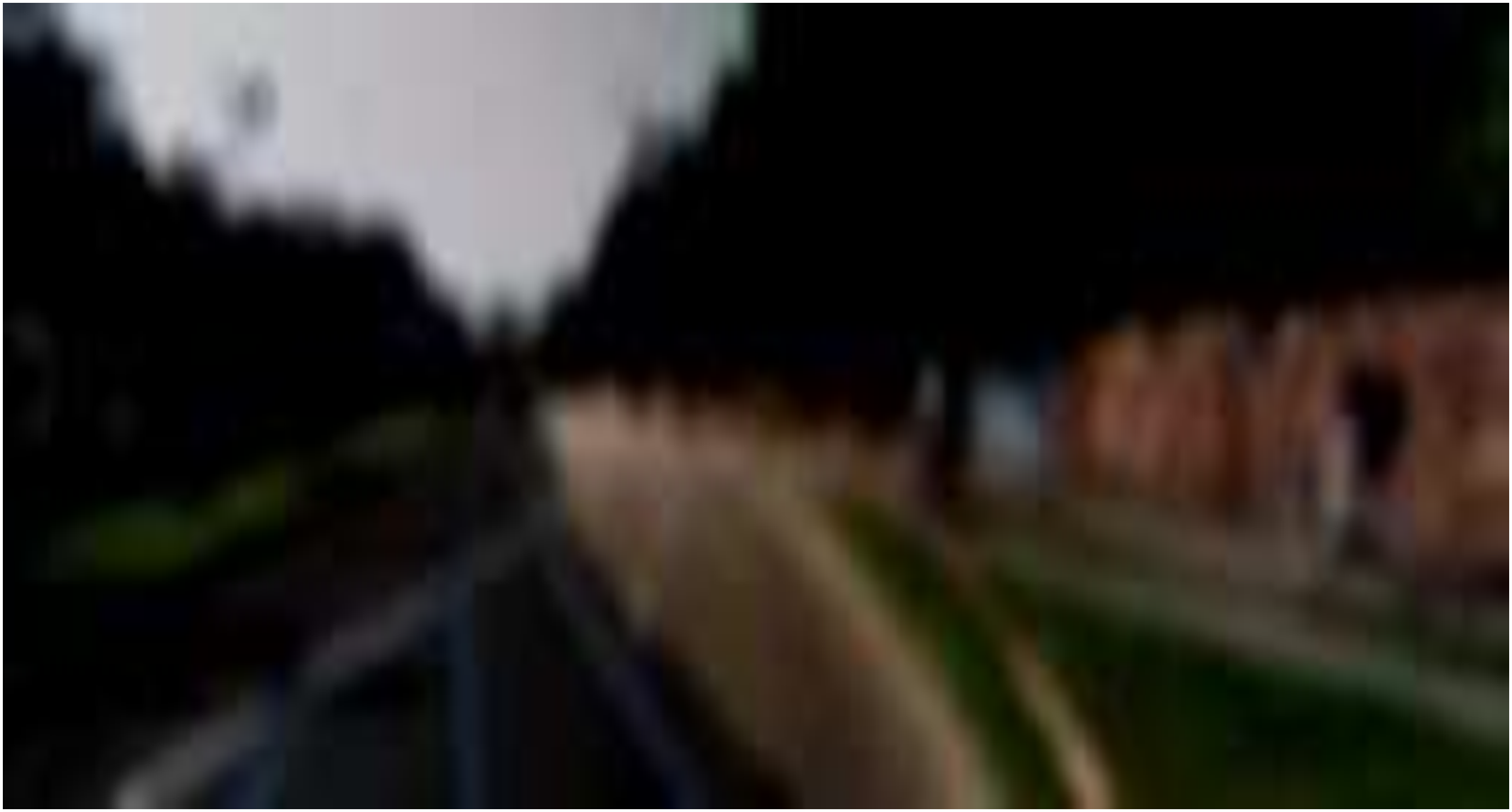}}
    {\vspace{-0.5em}\caption{\footnotesize Blurred image with radius = 3}\label{fig:vd_blurred_images_rad3}}
    \ffigbox[\FBwidth]
    {\includegraphics[height=\CommonHeight]{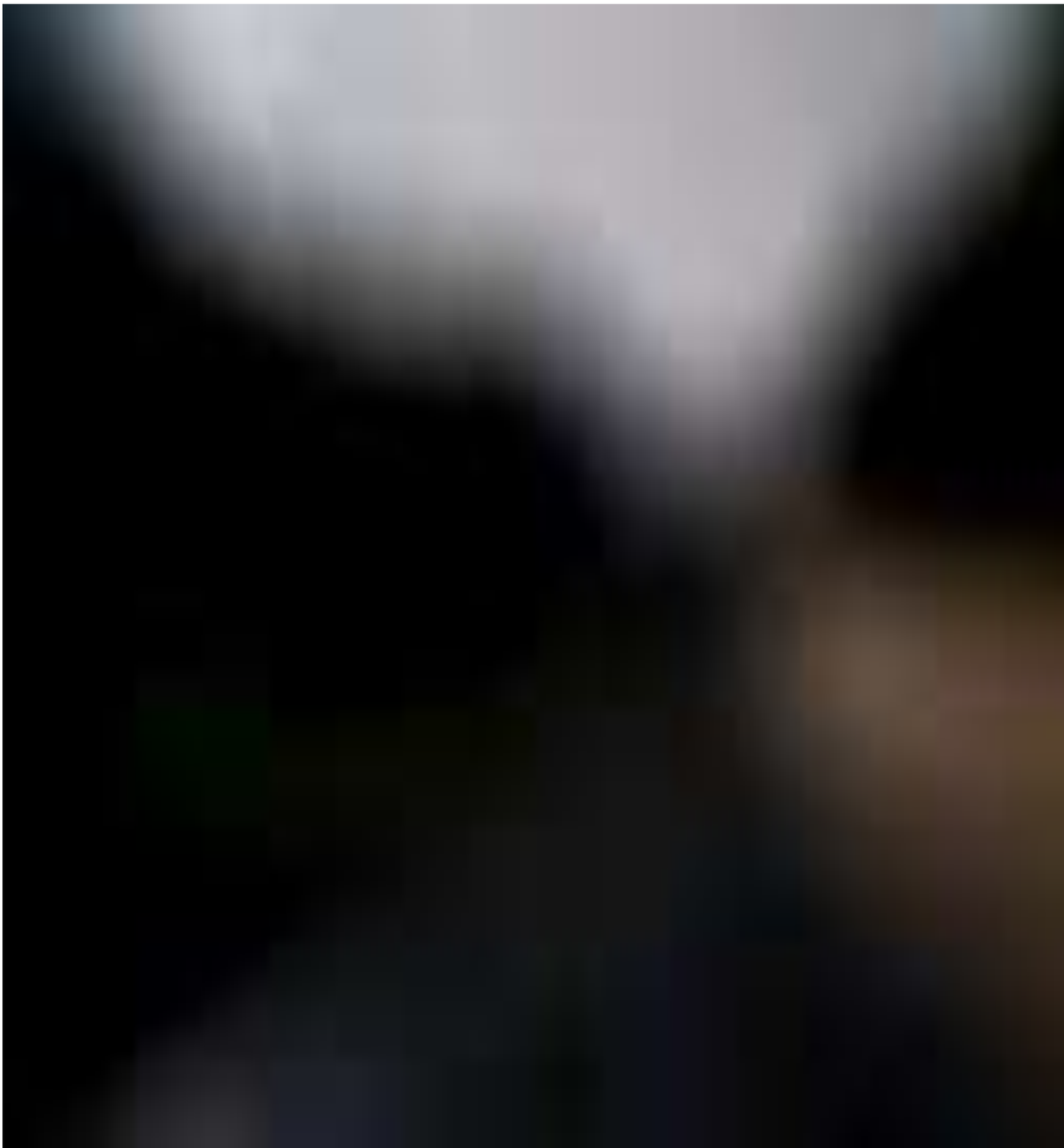}}
    {\vspace{-0.5em}\caption{\footnotesize Blurred image with radius = 10}\label{fig:vd_blurred_images_rad10}}
    \ffigbox[\FBwidth]
    {\includegraphics[height=\CommonHeight]{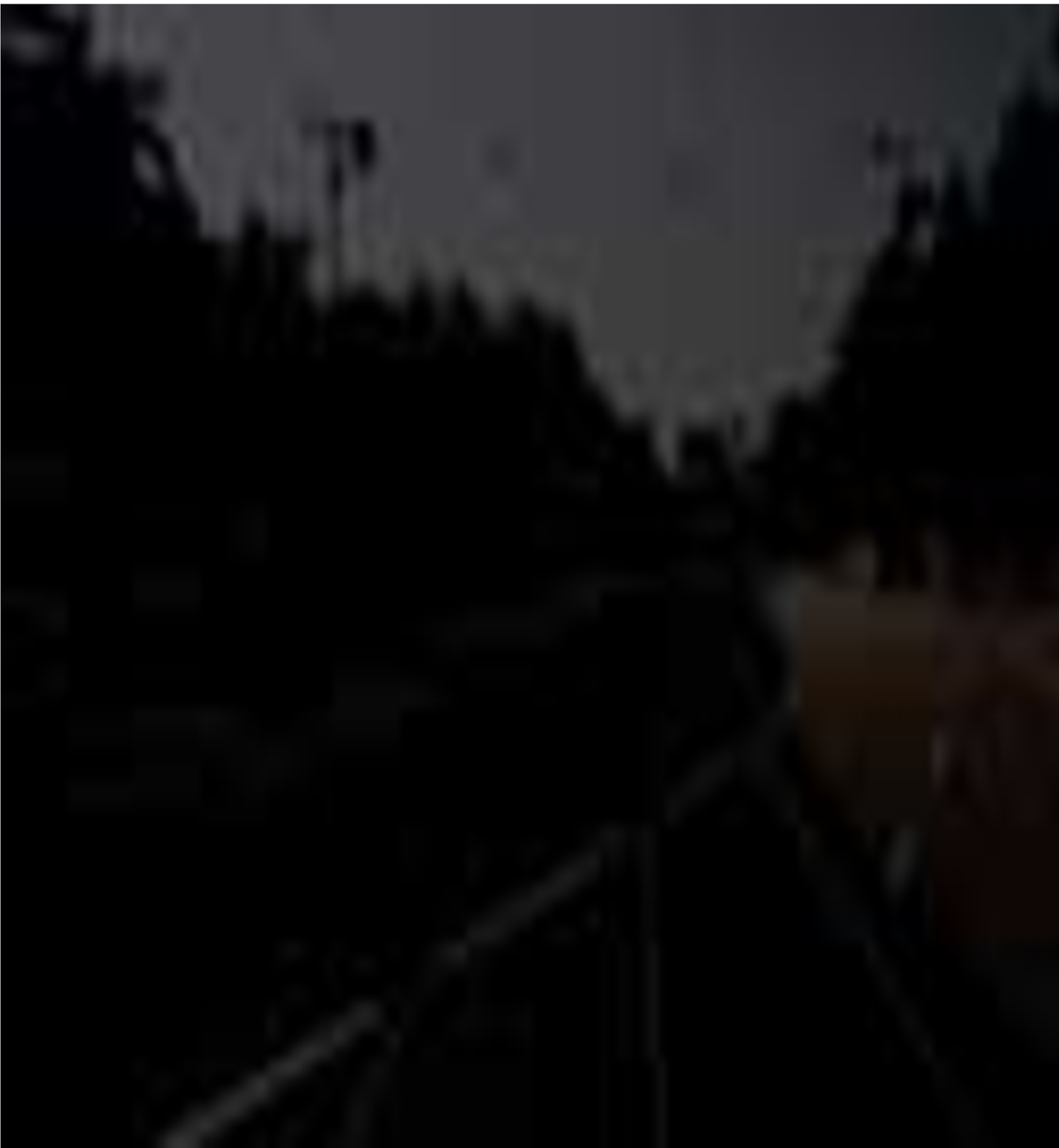}}
    {\vspace{-0.5em}\caption{\footnotesize Darkened image with max contrast = 0.4 and gamma = 1.5}\label{fig:vd_blurred_images_dark}}
      \end{subfloatrow}
    } 
    \vspace{-1.0em}
      \caption{\small Some artificial noise-added images as they have been tested in our experiments. \vspace{-1.3em}}  
       \label{fig:vd_blurred_images}
    
    }
\end{figure*}

\vspace{-1.0em}

\begin{table*}[ht!]
  \caption{\small Results obtained on KITTI dataset applying Gaussian blur to images and changing lighting conditions.}
  \label{tab:blur_exp}
  \begin{tabular}{|c||c|c|c|c|c|}
    \hline
    					& \small{Plain} & \small{Blur rad:3} & \small{Blur rad:10} & \small{Darkened} & \\
    \hline
    \small{thr. $\delta<1.25$} 		& 0.318 	& 0.244 & 0.142 & 0.176 & \small Higher\\
    \small{thr. $\delta<1.25^{2}$} 	& 0.617 	& 0.525 & 0.350 & 0.348 & \small is\\
    \small{thr. $\delta<1.25^{3}$} 	& 0.813 	& 0.741 & 0.573 & 0.509 & \small better\\
    \hline
    \small{RMSE} 					& 7.508 & 8.126 & 9.483 & 9.645 & \small Lower \\
    \small{Log RMSE} 				& 0.524	& 0.606 & 0.792 & 0.923 & \small is\\
    \small{Scale Inv.MSE} 			& 0.196	& 0.209 & 0.240 & 0.346 & \small better\\
    \hline
  \end{tabular}
\end{table*}

\subsection{Testing network robustness}

To test the robustness of our approach on different scene conditions, we performed additional experiments on the KITTI sequences, by  
producing transformed versions of each test sequence. To do so we changed contrast and gamma 
to simulate different light conditions, and applied Gaussian blur of different radius to simulate defocus or motion blur. 
In particular, we add a gaussian blur with a radius of 3 (we refer to this experiment as \textit{Blurred Image rad:3}, Figure \ref{fig:vd_blurred_images_rad3} ) and 10 
pixels (\textit{Blurred Image rad:10}, Figure \ref{fig:vd_blurred_images_rad10} ) and change image lighting by setting max contrast to 0.4 and gamma to 1.5 (\textit{Darkened Image}, 
Figure \ref{fig:vd_blurred_images_dark} ). 
The results of the evaluation with respect to these sequences are shown in Table \ref{tab:blur_exp}. It was not possible to test Eigen et al. method on blurred images since they did not publicly release their network's weights trained on KITTI dataset, so we compare our results with their performance on non-blurred images. 
On \textit{Blurred Image rad:3} experiment 
our performance is still better than Eigen et al. in terms of scale invariant log MSE error even after noise addition.  
and experience just a slight performance deterioration on other metric. On the \textit{Blurred Image rad:10} and \textit{Darkened Image} experiments, the estimations are less accurate, 
but results remains acceptable and comparable with other techniques on scale invariant log MSE metric. These experiments, thus, demonstrate our network capability to perform 
acceptable estimations even with very noisy images.

\section{Conclusion and Future Work} \label{sec:conclusions}

In this paper, we explore the architecture and performances of a depth estimation algorithm based on a Encoder-Decoder Convolutional Neural Networks architecture. The proposed algorithm is intended to be the foundation of an Obstacle Detection system, meant to be run by fast vehicles. We address the limitations of stereo systems using a learning approach trained on synthetic images with long range ground truth. We test two kind of inputs, monocular images and monocular images with optical flow. Both networks trained on synthetic data have shown, compared to state-of-the-art methods, good performances on real data, suggesting that this training strategy is able to overcome some weaknesses of learning approaches, such as generalization and training data availability. 
In addition, we showed how the proposed algorithm is capable of estimating depth even when the starting images have been corrupted with blur, darkened or lightened. 
In future work we plan to increase network's robustness augmenting virtual training data, also adding dynamical objects to the scene. In future works we will consider finetuning on real images to improve performance. In addition we plan to integrate our depth estimator together with semantic segmentation algorithms and object detectors to obtain a semantic knowledge of the scene, useful to infer information about estimation uncertainty and model more robust interpretations of the scene.
%

\bibliographystyle{IEEEtran}
\bibliography{bibliography}

\end{document}